\documentclass[11pt,a4paper]{article}

\usepackage[margin=1in]{geometry}
\usepackage{times}
\usepackage{graphicx}
\usepackage{booktabs}
\usepackage{hyperref}
\usepackage{url}
\usepackage{amsmath}
\usepackage{amssymb}
\usepackage{enumitem}
\usepackage[numbers,sort&compress]{natbib}
\usepackage{setspace}
\usepackage{float}
\usepackage{caption}
\usepackage{xcolor}

\hypersetup{
  colorlinks=true,
  linkcolor=blue!50!black,
  citecolor=blue!50!black,
  urlcolor=blue!50!black
}

\title{A Social Media Analysis of Discourse on the Israel--Palestine Conflict on Telegram}
\author{
Michail Zafeiropoulos$^{2}$
Despoina Antonakaki$^{1,2}$,
Sotiris Ioannidis$^{2}$ \\
\\
$^{1}$Institute of Computer Science, Foundation for Research and Technology,\\ Vassilika Vouton, Heraklion, Crete, Greece \\
$^{2}$Technical University of Crete, University Campus, \\Akrotiri, Chania, Greece \\
} 
\date{}

\begin{document}
\maketitle

\begin{abstract}
Social media has become a central arena in which armed conflicts are contested, yet the pro-Israel and pro-Palestine communities on Telegram, whose broadcast architecture yields an unusually direct record of deliberate political communication, have not been systematically compared at scale. This study presents a multi-method computational analysis of 87,617 messages from sixteen Telegram channels, eight pro-Israel and eight pro-Palestine, spanning May 2021 to June 2026 and covering multiple conflict escalations. It combines sentiment analysis, three stance detection methods drawn from distinct paradigms (keyword matching, zero-shot DeBERTa via natural language inference, and a fine-tuned BERTweet model), and a framing analysis, all evaluated against 736 manually annotated messages. The fine-tuned model performed best (72.1\% accuracy, 0.721 macro F1 under 5-fold cross-validation), outperforming both label-free baselines by 8 to 11 points; the baselines stalled in the low-to-mid 60s, indicating a hard ceiling for stance detection not adapted to in-domain language. The central finding emerges only when sentiment, stance, and framing are read together: the two communities deploy the same death- and victim-related vocabulary in opposite emotional registers, pro-Israel channels predominantly neutral and report-style, pro-Palestine channels markedly more negative, consistent with writing from the distinct discourse positions of acting party and affected party.
\end{abstract}


\section{Introduction}
\label{sec:introduction}

\subsection{Social Media and Armed Conflict}
\label{sec:sm-conflict}
The relationship between armed conflict and information has always been contested, but the rise of networked social media has transformed the scale, speed, and accessibility of that contest. Where political narratives once traveled through institutional channels, like governments, broadcasters, press agencies, they now flow simultaneously through millions of individual accounts, organizational channels, and algorithmic amplification systems operating in real time. Reuter et al. \cite{reuter2020} identify social media as a multidimensional communication infrastructure in conflicts and crises, operating across four distinct patterns: citizen-to-citizen coordination, authority-to-public crisis communication, citizen-generated content flowing back to authorities, and inter-organizational communication among institutions. Zeitzoff \cite{zeitzoff2017} demonstrates that conflict actors do not simply use social media to mobilize supporters but deploy it strategically to shape narratives for international audiences: during the 2012 Gaza conflict, both Israel and Hamas conducted parallel information campaigns on Twitter alongside their military operations, leading observers to describe it as the first "Twitter war." Oates \cite{oates2020} situates this within a structural argument about platform economics: because social media companies profit by selling engaged audiences to advertisers, and because emotionally charged content drives that engagement, effective content moderation runs directly against their commercial interests, leaving platforms permanently hospitable to unverified, emotionally resonant political content regardless of its accuracy. The consequences for information quality in conflict contexts are well-documented and severe. Vosoughi, Roy, and Aral \cite{vosoughi2018} analyzed approximately 126,000 news stories spread on Twitter from 2006 to 2017 and found that false news spreads significantly farther, faster, deeper, and more broadly than accurate information across all content categories, with false political news exhibiting the most pronounced diffusion dynamics. This asymmetry was not attributable to automated bots but to human users who preferentially shared novel, emotionally resonant content regardless of veracity. Automated accounts nonetheless play a distinct amplifying role: analysis of the 2020 U.S. presidential election found that bots were substantially over-represented among accounts spreading conspiratorial content and were used to inflate the apparent reach of partisan narratives, distorting which messages appear to command genuine public support \cite{ferrara2020}. In conflict environments, where ground-level information is difficult to verify and emotional stakes are high, these dynamics allow unverified claims about casualties, atrocities, and military operations to achieve global circulation ahead of any correction. Haq et al. \cite{haq2022} extend this analysis to the structural dimension: platforms produce information divides, polarized communities in which algorithmically reinforced echo chambers limit cross-cutting exposure, that are systematically exploited by actors engaged in information warfare. Oakley and Rogg \cite{oakley2024} further demonstrate that social media blurs the boundary between civilian observer and conflict participant, enabling geographically distant users to influence military and political dynamics while simultaneously complicating the intelligence picture for professional analysts. Bennett and Livingston \cite{bennett2018} locate these developments within a broader account of democratic disruption, arguing that the breakdown of authoritative information flows between institutions and publics constitutes a structural condition of contemporary political communication rather than an isolated pathology.

\subsection{Telegram as Political Communication Infrastructure}
\label{sec:telegram-infra}
Among the platforms through which conflict-related political communication now flows, Telegram occupies a distinctive and analytically important position. The platform's emergence as infrastructure for politically motivated actors predates the Israel-Palestine conflict's digital escalation. Prucha \cite{prucha2016} documented its systematic adoption by jihadist movements following Twitter's increasingly effective content moderation, observing that when Twitter suppressed IS networks, those networks migrated en masse to Telegram. The architectural features driving this migration were specific: Telegram offered encryption, no limits on data sharing, and, crucially, resistance to content removal, since channel content could be forwarded easily to private channels. These properties established Telegram as moderation-resistant broadcast infrastructure, a stable environment for actors whose content faced suppression elsewhere. Unlike algorithmically curated platforms, Telegram's channel architecture allows administrators to broadcast content directly and without interruption to arbitrarily large subscriber audiences, without editorial filtering or engagement-based ranking. Channel content therefore directly reflects what administrators choose to publish, making it an unusually transparent record of deliberate political communication. Antonakaki and Ioannidis \cite{antonakaki2025telegram} document how these same structural features shaped the conflict's digital communication after October 7, 2023, with Telegram serving as a key platform for real-time information dissemination, propaganda, and audience mobilization across both pro-Palestinian and pro-Israeli communities. Both sides used the platform to distribute updates, frame events, and project political narratives to international audiences, though its reach was repeatedly constrained by content-moderation and censorship efforts targeting affiliated channels.

\subsection{The Israel-Palestine Conflict as a Case Study}
\label{sec:case-study}
The Israel-Palestine conflict represents a particularly significant case study for digital communication research. It is among the most extensively discussed political topics on social media globally, generating persistent high-volume discourse that predates the October 2023 escalation and has intensified through subsequent events. Osimen et al. \cite{osimen2025} document how the conflict produced asymmetric media framing from the earliest phases of the current escalation, with mainstream Western outlets predominantly adopting Israeli narratives and framing in the immediate aftermath of October 7, while Palestinian perspectives and the broader historical context received comparatively limited coverage. Earlier empirical work established Twitter as a central arena for this contest: Siapera, Hunt, and Lynn \cite{siapera2015} analysed roughly three million tweets from the 2014 Gaza war and found a discourse dominated by activist and witness accounts, a pronounced pro-Palestinian volume, and a recurring humanitarian frame cutting across both partisan poles \cite{siapera2015}. Social media platforms, and Telegram in particular, became arenas in which communities contested and supplemented this mainstream framing. Antonakaki and Ioannidis \cite{antonakaki2025telegram} trace this dynamic across Telegram, Twitter, and Reddit, identifying distinct discourse communities on each platform with different sentiment profiles and topical emphases. The conflict's global resonance, the clear ideological alignment of its most active online advocates, and the availability of large-volume Telegram data across an extended time period make it both an important subject of study in its own right and a productive case for examining how opposing communities construct and sustain political narratives through digital communication infrastructure.

\subsection{Motivation}
\label{sec:motivation}

Despite growing scholarly attention to conflict-related social media discourse, systematic large-scale computational analyses comparing pro-Israel and pro-Palestine Telegram communities remain limited. Existing studies have either examined the conflict through cross-platform analyses, where Telegram is considered alongside other social media platforms \cite{antonakaki2025telegram}, or have focused exclusively on platforms such as Twitter/X. To the best of our knowledge, no previous work has conducted a dedicated longitudinal NLP analysis comparing ideologically opposed Telegram communities using multiple complementary methods for sentiment analysis, stance detection, and narrative framing.

\subsection{Contributions}
\label{sec:contributions}

The main contributions of this work are summarized as follows:

\begin{itemize}
    \item We present, to the best of our knowledge, the first large-scale computational analysis comparing pro-Israel and pro-Palestine Telegram channels. The study is based on a corpus of 87,617 messages collected from 16 public Telegram channels spanning May~2021 to June~2026 and covering multiple phases of the Israel--Palestine conflict.

    \item We construct and release a curated dataset comprising Telegram messages annotated for stance, enabling the systematic evaluation of supervised and zero-shot stance detection methods on conflict-related social media content.

    \item We perform a comprehensive empirical comparison of three stance detection approaches, namely keyword matching, zero-shot DeBERTa inference, and a fine-tuned BERTweet classifier. The proposed fine-tuned model achieves an accuracy of 72.1\% and a macro F1-score of 0.721 under five-fold cross-validation.

    \item We integrate sentiment analysis, stance detection, and narrative framing into a unified analytical framework, allowing the evolution of online discourse to be examined in relation to major conflict events over time.

    \item We provide a longitudinal analysis of messaging strategies across opposing communities, highlighting differences in sentiment, stance, and framing and how these evolve during periods of military escalation.
\end{itemize}

\subsection{Study Structure}
\label{sec:study-structure}
The study is organized as follows. Section~\ref{sec:background} provides the theoretical background. Section~\ref{sec:related} reviews related work. Section~\ref{sec:dataset} describes the dataset and preprocessing pipeline. Section~\ref{sec:methodology} presents the methodology. Section~\ref{sec:results} reports empirical results. Section~\ref{sec:discussion} discusses findings in relation to the research questions. Section~\ref{sec:conclusion} concludes.

\section{Theoretical and Computational Background}
\label{sec:background}

\subsection{Social Media and Political Communication}
\label{sec:sm-polcomm}
The emergence of social media has restructured the architecture of political communication in ways that extend well beyond the digitization of existing practices. Social media, defined as internet-based applications, built on the ideological and technological foundations of Web 2.0, that enable the creation and exchange of user-generated content \cite{kaplan2010}, have become increasingly central to shaping political discourse globally. Where political communication once depended on institutionally controlled channels, governments, broadcasters, editorial gatekeepers, social media now allow state actors, advocacy organizations, and ordinary users to broadcast political content directly to mass audiences. This shift has structural consequences for how political agendas are formed, how events are framed, and how communities of political interest organize and sustain themselves. Agenda-setting theory offers one of the most durable frameworks for understanding how media shape political perception. McCombs and Shaw's foundational work established that mass media do not tell audiences what to think, but rather what to think about, by conferring salience on certain issues over others, demonstrating a correlation of +.967 between media emphasis on campaign issues and voters' independent judgments of issue importance \cite{mccombs1972}. In the social media environment this process has been modified but not dissolved. Feezell demonstrates experimentally that exposure to political information through social media increases the probability that recipients will identify covered issues as important, and that this effect is strongest among users with low political interest, precisely those who would not otherwise seek out news \cite{feezell2018}. Social media thus extends the reach of agenda-setting through incidental exposure: users who engage with platforms primarily for social purposes encounter politically relevant content within the same feed, without actively choosing to consume it. This mechanism is directly relevant to Telegram channel subscribers, who receive conflict-related content continuously regardless of whether political engagement was their primary motivation for using the platform. Framing theory provides the analytical vocabulary for understanding how content shapes interpretation once it reaches audiences. Entman defines framing as the selection and salience of aspects of perceived reality in a communicating text, in a way that promotes a particular problem definition, causal interpretation, moral evaluation, or treatment recommendation \cite{entman1993}. Frames do not merely describe events; they structure how events are understood by determining which elements are foregrounded and which are suppressed. Scheufele elaborates a distinction between agenda-setting and frame-setting, arguing that while the former concerns the salience of issues, the latter concerns the salience of issue attributes, the specific dimensions through which an issue is interpreted \cite{scheufele1999}. The concept originates in Goffman's broader sociological account of how individuals organize experience through interpretive schemata (Goffman, 1974, as cited in Scheufele \cite{scheufele1999}). In politically polarized conflict communication, opposing communities apply competing frames to identical events: the same military operation can be simultaneously framed as a legitimate counter-terrorism measure or a war crime depending on which attributes are made salient. The framing analysis conducted in Section~\ref{sec:methodology} directly operationalizes this theoretical claim. Alternative media ecosystems have become significant channels through which competing frames circulate outside mainstream institutional contexts. Haller, Holt and de la Brosse characterize alternative media as outlets run by actors who publish interpretations of current events in direct response to the perception that their political or ideological perspective is being treated unfairly by mainstream media \cite{haller2019}. This self-positioning as a corrective to perceived bias characterizes many of the Telegram channels in this dataset, which frame their content explicitly against what they present as distorted mainstream coverage. Fenton and Barassi caution that techno-optimistic accounts of social media as straightforwardly empowering tools for political movements overlook how these platforms embed political communication within commercial architectures that are not neutral \cite{fenton2011}. Leung and Lee demonstrate empirically that internet alternative media usage is associated with higher rates of political participation and with more critical attitudes toward mainstream media, consistent with the view that alternative media consumption is bound up with active political engagement rather than passive reception \cite{leung2014}. Evidence on echo chambers is mixed and depends heavily on the environment studied. In open online-news settings, the strong version of the study is contested. Garrett \cite{garrett2009} finds that while users prefer opinion-reinforcing information, their avoidance of opinion-challenging content is weak, and they remain exposed to opposing views. Messing and Westwood similarly demonstrate that social endorsement cues, signals that content has been recommended by others in one's network, can override partisan source preferences and reshape information exposure, with the mere presence of such endorsements reducing partisan selectivity to levels indistinguishable from chance \cite{messing2014}. Structural echo chambers emerge, however, where platform architecture actively promotes homophily. Cinelli et al. \cite{cinelli2021} show that homophilic clustering and like-minded information diffusion dominate on platforms organised around social networks and feed algorithms (Facebook, Twitter), while remaining largely absent on others (Reddit, Gab). The channels analysed here represent a limiting case: a one-directional broadcast architecture in which subscribers self-select into a single political stream with no mechanism, algorithmic or social, delivering cross-cutting content. In this environment the conditions Garrett found insufficient to produce echo chambers in open browsing are structurally guaranteed, and the endorsement-based corrective identified by Messing and Westwood has no channel through which to operate. Baumann et al. \cite{baumann2020} model the dynamics through which opinion reinforcement and homophilic interaction jointly produce polarized opinion distributions in networked environments, identifying the controversialness of a topic as a key driver of radicalization dynamics. Read together, these frameworks describe an information environment in which politically aligned channels can maintain coherent, self-reinforcing frames while reaching large audiences predisposed to receive them. This environment is precisely what Telegram's channel architecture instantiates, and it constitutes the theoretical context within which the corpus analyzed in this study must be understood.

\subsection{Information Warfare and Conflict Communication}
\label{sec:infowar}
The relationship between information and armed conflict predates the digital era, but the emergence of social media has dramatically expanded the scale, speed, and accessibility of information operations. Social media platforms have been widely characterized as vulnerable to manipulation and abuse, from automated astroturfing and propaganda to coordinated misinformation campaigns that exploit the speed and reach of networked communication \cite{ferrara2015}. Libicki's foundational taxonomy identifies seven distinct forms of information warfare, including psychological warfare, command-and-control warfare, and electronic warfare, each involving the protection, manipulation, degradation, or denial of information \cite{libicki1995}. While these categories reflect Cold War-era strategic thinking, the underlying logic, that controlling the information environment confers strategic advantage, has intensified with the shift to networked digital communication. What has changed is not the instrumental logic of information warfare but the infrastructure through which it is conducted. Strategic narrative theory offers the most developed framework for understanding how belligerents construct and project meaning during conflict. Roselle, Miskimmon and O'Loughlin argue that the study of strategic narrative is central to understanding how all aspects of a conflict are defined, constructed and understood, with combatants' grievances, claims, and aspirations subject to characterization, the attribution of motives, and the construction of reputation \cite{roselle2014}. Strategic narratives operate simultaneously at multiple levels: system narratives that situate a conflict within broad geopolitical frameworks, national narratives that define a state's identity and historical role, and issue narratives that frame specific events within those larger structures. Roselle et al. further observe that digitization disrupts the sequential structure of conflict narratives, creating temporal fragility in which the apparently settled meaning of past events can be destabilized by the emergence of new data or images that force their reconsideration \cite{roselle2014}. This fragility is a defining feature of the information environment around the Israel-Palestine conflict, where competing narratives about identical events are continuously contested across platforms in real time. Lewandowsky et al. situate this dynamic within a broader account of how narratives function cognitively in conflict contexts: parties involved in a conflict nearly always create a conflict-supporting narrative that provides explanation and justification for their involvement, buttressing beliefs about the justness of one's own cause while delegitimizing the opponent \cite{lewandowsky2013}. Crucially, frames and narratives are not equivalent to propaganda, they are, as Lewandowsky et al. argue, necessary cognitive tools designed to pare down information in order to manage complexity \cite{lewandowsky2013}. What makes conflict narratives politically consequential is their resilience: misinformation embedded in a dominant narrative can persist for years after factual correction, as demonstrated by the sustained prevalence of false WMD beliefs in the American public following the Iraq War \cite{lewandowsky2013}. In politically polarized conflict discourse, where opposing communities inhabit entirely separate information environments, this resistance to correction is structurally amplified. The digital transformation of conflict communication has introduced new operational forms alongside traditional propaganda. Woolley and Howard define computational propaganda as an emergent form of political manipulation that occurs over the internet, the assemblage of social media platforms, autonomous agents, algorithms, and big data tasked with the manipulation of public opinion \cite{woolley2018}. Haq et al. document how social media platforms are systematically exploited to create information divides, algorithmically reinforced echo chambers that limit cross-cutting exposure and can be weaponized to sustain parallel, incompatible narratives of the same conflict \cite{haq2022}. Oakley and Rogg extend this analysis to intelligence contexts, arguing that the proliferation of propaganda and open-source content on social media produces a smog of war that complicates the information picture for analysts and civilians alike \cite{oakley2024}. Digital diplomacy constitutes a further dimension of conflict communication in the social media era. Manor and Crilley's \cite{manor2018} analysis of the Israeli Ministry of Foreign Affairs' Twitter activity during the 2014 Gaza War demonstrates how state actors use social media to craft and disseminate frames directly to connected audiences, circumventing traditional media gatekeepers \cite{manor2018}. Their analysis identifies fourteen distinct frames deployed across the conflict period, each containing the four elements defined by Entman, problem definition, causal attribution, moral evaluation, and treatment recommendation, and each designed to legitimize Israeli policy while delegitimizing Hamas. Manor and Crilley further observe that framing in conflict is simultaneously proactive and reactive, as actors must respond to the framing of opponents and press coverage while maintaining narrative coherence \cite{manor2018}. Zeitzoff's broader analysis confirms that conflict actors use social media not merely to report events but to strategically shape international audience perceptions and journalistic coverage, as demonstrated during the 2012 Gaza conflict \cite{zeitzoff2017}. In combination, these frameworks position conflict communication on social media not as a distorted reflection of events on the ground but as a domain of active contestation in which narrative construction and strategic framing are integral to how conflicts are fought and understood. The Telegram channels analyzed in this study are participants in precisely this form of information contest: each channel constructs frames that legitimize one side's position and delegitimize the other's, disseminating those frames to large subscriber audiences in real time. Analyzing these channels through sentiment, stance, and framing provides a systematic window into how opposing communities sustain competing conflict narratives across an extended period.

\subsection{Telegram as a Political Communication Platform}
\label{sec:telegram-platform}
Telegram is a cloud-based instant messaging service offering both private messaging and public broadcasting capabilities. It reached one billion messages per day within sixteen months of operation and has since grown into one of the largest messaging platforms globally \cite{dargahi2017}. Its rise as infrastructure for politically motivated communication is not incidental but structural: the platform's architecture offers a set of features that distinguish it sharply from algorithmically curated social media networks and make it particularly suited to the large-scale, unfiltered distribution of political content. The platform supports three distinct communication modes. User-to-user private messaging operates similarly to other messaging services. Groups allow many-to-many communication in which all members can send and receive messages. Channels operate differently: they are broadcasting mechanisms in which one or a small number of administrators publish content to an unlimited subscriber audience, with no ability for subscribers to respond within the channel itself \cite{dargahi2017}. Unlike Facebook, Twitter, and other major social networks, there is no social graph or friendship relation between users in Telegram; channels reach subscribers directly without intermediary social connections \cite{dargahi2017}. Critically, Telegram channels present posts to subscribers in chronological order, with no engagement-based ranking or algorithmic recommendation applied to channel content: what an administrator publishes is what subscribers receive, in the order it was posted. This makes channel content an unusually direct record of deliberate communication choices by the channel operators rather than a product of platform optimization. Public channel content is additionally not searchable via general search engines, limiting discovery to users who already possess a channel link or username \cite{dargahi2017}, which reinforces the self-contained character of Telegram communities. A further architectural feature has shaped Telegram's adoption by politically motivated actors: its consistent resistance to content removal compared to other platforms. When channels face suspension, their content can be forwarded easily to private channels, maintaining the continuity of content networks even under moderation pressure \cite{prucha2016}. Combined with optional end-to-end encryption for private communications, these features established Telegram as moderation-resistant broadcast infrastructure, a stable environment for actors whose content faced suppression elsewhere. These properties drove the systematic adoption of Telegram by jihadist networks from 2016 onwards, following Twitter's increasingly effective content moderation. Prucha documents how IS media operations migrated from Twitter to Telegram as suppression intensified: while the Twitter ship was sinking for IS, the jihadi online swarm simply turned to a new social media platform, Telegram \cite{prucha2016}. The migration was structurally motivated: Telegram offered encryption, large file sharing, resistance to deletion, and channel forwarding capabilities that Twitter's open, moderated environment could not match \cite{prucha2016}. Prucha's analysis establishes a precedent that extends well beyond jihadist movements, Telegram's architecture makes it attractive to any actor whose content faces institutional pressure or who requires direct, unmediated access to a large audience. Governments, political movements, news organizations, and conflict actors across the political spectrum have since incorporated Telegram into their communication infrastructures for precisely these reasons. In the Israel-Hamas conflict, Telegram became a key platform for real-time information dissemination on both sides. Antonakaki and Ioannidis document a measurable increase in Telegram use among both Israelis and Palestinians following October 7, 2023, with the platform becoming a prominent channel through which information, propaganda, and live updates were distributed throughout the conflict \cite{antonakaki2025telegram}. The same study notes that participants and observers used Telegram specifically because of its encrypted communication given the restricted conditions of the conflict zone \cite{antonakaki2025telegram}. A subsequent cross-platform analysis characterizes Telegram as a hub for uncensored documentation during the conflict, with its encrypted channels enabling real-time, unfiltered distribution of events as they developed \cite{antonakaki2026crossplatform}. The dataset analyzed in this study was collected entirely from Telegram channels for these structural reasons. The channel architecture produces a communication record that directly reflects what administrators chose to publish, unmediated by algorithmic ranking or engagement-based selection. Sentiment, stance, and framing patterns in the data therefore reflect the editorial decisions of channel operators rather than platform optimization processes. The following chapter describes the specific channels selected, the collection procedure, and the resulting dataset's composition.

\subsection{Sentiment Analysis}
\label{sec:sentiment-bg}

\subsubsection{Definition and Scope}
\label{sec:sentiment-def}
Sentiment analysis is the computational task of identifying the subjective orientation expressed in natural language text, typically classified as positive, negative, or neutral \cite{taboada2011}. The task can be applied at multiple levels of granularity: document-level classification assigns a single polarity label to an entire text, sentence-level analysis classifies individual statements, and aspect-level analysis attributes sentiment to specific entities or topics mentioned within a text. In political communication research, document- and sentence-level analysis are most common. A critical limitation of sentiment analysis in political contexts is that emotional valence and political positioning are not the same thing: a message can carry strong negative sentiment while expressing support for a political actor, or describe an adversary's defeat in a neutral factual register.
This distinction between emotional tone and political stance motivates the complementary methods described in Section~\ref{sec:stance}.

\subsubsection{Lexicon-Based Approaches}
\label{sec:sentiment-lexicon}
The earliest systematic computational approaches to sentiment analysis were lexicon-based. These methods rest on the assumption, formalized in psycholinguistic research and adopted widely in the NLP literature, that individual words carry prior polarity, a semantic orientation that is independent of context and expressible as a numerical value \cite{taboada2011}. The Semantic Orientation Calculator (SO-CAL), described by Taboada et al., operationalizes this assumption through manually constructed dictionaries covering adjectives, nouns, verbs, and adverbs, each word assigned a hand-ranked score on a scale from +5 to -5 \cite{taboada2011}. The overall sentiment of a document is computed by summing the scores of the relevant words it contains. SO-CAL extends beyond simple word scoring to handle two important linguistic phenomena. First, intensification: amplifiers such as very or extraordinarily multiply the score of an adjacent lexical item upward, while downtoners such as slightly or somewhat reduce it, each assigned a percentage modifier reflecting their relative strength \cite{taboada2011}. Second, negation: the system shifts the polarity of a lexical item toward the opposite value when a negating expression is found in its context window, including not, never, nobody, and equivalent constructions such as without \cite{taboada2011}. The advantage of lexicon-based methods is full interpretability and no requirement for labeled training data. Their fundamental limitation is coverage and context-insensitivity: words absent from the lexicon contribute nothing, and identical words are scored identically regardless of syntactic structure or surrounding meaning.

\subsubsection{Machine Learning Approaches}
\label{sec:sentiment-ml}
Supervised machine learning approaches treat sentiment classification as a standard text categorization problem, distinguishing two "topics", positive and negative sentiment, rather than thematic categories \cite{pang2002}. Input text is converted into a numerical feature vector, typically using bag-of-words counts or TF-IDF weightings, and a classifier is trained on labeled examples to map these representations to polarity labels. Naïve Bayes, maximum entropy (logistic regression), and support vector machines were the dominant algorithms in this paradigm \cite{pang2002}. These methods learn statistical associations between surface features and sentiment labels that lexicons cannot capture. Their feature representations are nonetheless insensitive to word order and syntactic structure, which makes negation and its scope difficult to resolve: a bag-of-words model struggles to recognize that negating "The movie was terrible" yields "The movie was not terrible," a phrase whose sentiment is less negative rather than simply reversed \cite{socher2013}. This compositionality problem limits performance on texts where sentiment depends on phrase- or sentence-level structure, which is common in political messaging.

\subsubsection{Transformer-Based Approaches}
\label{sec:sentiment-transformer}
The Transformer and Self-Attention The transformer architecture introduced by Vaswani et al. \cite{vaswani2017} replaced sequential processing with self-attention, a mechanism in which every token attends directly to every other token in the sequence, regardless of distance. Each token's representation is updated as a weighted combination of all the others, with the weights determined by learned query--key projections \cite{vaswani2017}, \cite{deberta2021}. This direct modelling of all pairwise relations is what distinguishes transformers from earlier recurrent architectures, which process tokens sequentially and struggle to propagate information across long distances. In practice the operation is run in parallel as multi-head attention, allowing different heads to capture distinct syntactic, semantic, and positional relationships at once. BERT BERT (Bidirectional Encoder Representations from Transformers) \cite{devlin2019} applied the transformer encoder to general-purpose language understanding through large-scale pre-training. Its central contribution is bidirectionality: through a masked language-modelling objective, in which a fraction of input tokens are hidden and predicted from both left and right context, every layer builds representations that encode the full sentential context of each word rather than the left-to-right history of earlier models. A special [CLS] token provides an aggregate sequence representation, and for downstream tasks the model is fine-tuned by appending a classification layer and updating all parameters on labeled data \cite{devlin2019}. Because the pre-trained weights already encode rich linguistic structure, only modest amounts of labeled data are needed to adapt BERT to a new task, the property the fine-tuned stance model in this study relies on. RoBERTa RoBERTa \cite{liu2019} left BERT's architecture unchanged but showed it had been significantly undertrained: with dynamic rather than static masking, removal of the next-sentence-prediction objective, larger batches, and roughly ten times more training data (160GB vs. 16GB), it improved consistently across standard benchmarks. The result established that much of what had been credited to model design was in fact a matter of training procedure, a finding directly relevant to the tweet-adapted models described next, which apply the same recipe to social-media text. Twitter-RoBERTa and TweetEval Neither BERT nor RoBERTa was pre-trained on social media text, and this matters for a corpus of Telegram messages. Barbieri et al. \cite{barbieri2020} establish through the TweetEval benchmark that social media language presents specific challenges: an informal, conversational register, platform constraints such as character limits, and limited contextual cues in short texts differ systematically from Wikipedia and news text. A RoBERTa model additionally pre-trained on 60 million tweets (the RoB-RT approach in Barbieri et al.) consistently outperforms the base model on tweet classification tasks \cite{barbieri2020}. The cardiffnlp/twitter-roberta-base-sentiment-latest model used in this study follows this direction: it is a RoBERTa model that was further pre-trained on Twitter data and fine-tuned specifically on the TweetEval sentiment analysis task, producing a model whose pre-training and fine-tuning distributions both align with short, informal, politically charged social media content.

\subsubsection{Applications in Political Communication Research}
\label{sec:sentiment-apps}
Sentiment analysis has been applied to political communication contexts including electoral social media, protest movements, and conflict-related discourse. A consistent finding across these applications is that aggregate sentiment distributions in political corpora are dominated by neutral content even in explicitly activist sources, because a large share of political social media consists of event reporting, framing, and institutional updates rather than direct emotional expression. This is consistent with the distribution observed in the present corpus (Section~\ref{sec:results-sentiment}): approximately 64\% of messages were classified as neutral. This result illustrates why sentiment alone is analytically insufficient for politically aligned content; it captures emotional register but not political direction, which requires the stance detection methods described in Section~\ref{sec:stance}.

\subsection{Stance Detection}
\label{sec:stance}
\subsubsection{The Distinction Between Sentiment and Stance}
\label{sec:stance-distinction}
Sentiment analysis and stance detection are related but fundamentally different tasks. Sentiment analysis asks what emotional valence a text expresses. Stance detection asks what position the author takes toward a specified target. Küçük and Can identify the critical divergence: the polarity of the text may be positive while the stance may be against a particular target, and vice versa \cite{kucuk2020}. A message expressing grief over civilian deaths carries negative sentiment; whether it constitutes support for Palestine or mourning for Israeli hostages depends on stance, not emotional valence. Sentiment classifiers cannot resolve this ambiguity. In a corpus drawn from politically aligned conflict channels, where the same events are described from opposing political positions, applying sentiment analysis alone systematically fails to capture political meaning. Stance is defined by Küçük and Can as the position of the text producer toward a target or set of targets \cite{kucuk2020}. The standard label set is \{Favor, Against, Neither\}, where the target, an entity, event, claim, or political position, must be specified in advance. The Neither class requires careful interpretation: Küçük and Can distinguish it from genuine neutrality, noting that if the stance is not Favor or Against, the author's stance may not be necessarily Neutral, but instead no stance information can be extracted from the text alone \cite{kucuk2020}. In a corpus drawn entirely from politically aligned channels, the absence of explicit stance markers in a message does not indicate neutrality; it indicates indirect framing.

\subsubsection{Target-Dependent Stance Detection}
\label{sec:stance-target}
Stance is not a property of text in isolation but a relation between text and target. The stance target may not be explicitly given in the input text \cite{kucuk2020}, requiring a model to infer political positioning from indirect signals, vocabulary selection, actor references, framing choices, rather than surface-level keywords or explicit statements. This is the property that makes stance detection substantially harder than sentiment classification and that makes simple lexicon matching insufficient for texts where stance is expressed through what is left unsaid, through metaphor, or through the selection of which actors to foreground. In this study, the target is fixed as the Israel--Palestine conflict, and each message is assigned one of three stances toward it, pro-Palestine, pro-Israel, or neutral, at the message level. Stance labels across the full corpus are produced by the three detection methods, while a separate set of 736 messages was labeled by hand to serve as ground truth for training and evaluation (Section~\ref{sec:annotation}). Because stance is judged per message rather than inherited from the channel, a message can be labeled neutral even when its channel is politically aligned, for instance, factual or report-style messages that describe an event without explicit stance markers, or that quote or report opposing viewpoints.

\subsubsection{Feature-Based and Deep Learning Approaches}
\label{sec:stance-features}
Feature-based machine learning approaches dominated early stance detection research. Küçük and Can document that SVM is by far the most commonly employed approach, appearing in more than 40 stance detection studies as either the primary method or baseline \cite{kucuk2020}. These systems rely on lexical features such as bag-of-words, character n-grams, and stance-indicative word lists, combined with sentiment lexicon scores and interaction features such as retweet patterns \cite{kucuk2020}. Mohammad et al.'s SemEval-2016 shared task \cite{mohammad2016}, the foundational stance detection benchmark, covering five targets including climate change, feminism, and Hillary Clinton, used an SVM baseline that outperformed all participating systems, demonstrating the continued competitiveness of carefully engineered lexical features when annotated data is limited. Deep learning approaches subsequently became standard. LSTM-based models are the most common neural architecture in the stance literature, and the introduction of attention mechanisms, allowing the model to dynamically weight tokens differently when producing a classification, is reported to improve performance consistently across multiple studies \cite{kucuk2020}. More recently, transformer-based models pre-trained on large corpora have displaced earlier architectures through transfer learning, adapting general-purpose representations to stance classification with modest amounts of labeled data.

\subsubsection{DeBERTa and Zero-Shot Stance via Natural Language Inference}
\label{sec:stance-deberta}
The zero-shot stance detection method applied in this study uses the MoritzLaurer/deberta-v3-large-zeroshot-v2.0 model, built on the DeBERTa architecture developed by He et al. \cite{deberta2021}. Understanding why DeBERTa is well-suited to this task requires understanding both its architectural innovation and the NLI framing of zero-shot classification. DeBERTa improves on BERT and RoBERTa through two innovations \cite{deberta2021}. First, disentangled attention represents each token with separate content and relative-position vectors rather than summing them, so that the attention between two words depends jointly on what they are and how far apart they sit, motivated by the observation that the dependency between words is not a function of their content alone. Second, an enhanced mask decoder reintroduces absolute-position information just before the output layer rather than at the input. Together these yield strong natural-language-inference performance: DeBERTa-large surpasses RoBERTa-large on MNLI while training on roughly half the data \cite{deberta2021}. That NLI strength is precisely what the zero-shot stance method exploits; the checkpoint used here is the later DeBERTa-v3 variant, which adds ELECTRA-style pre-training. Zero-shot stance detection exploits DeBERTa's strong NLI performance. NLI is the task of determining whether a premise text entails, contradicts, or is neutral toward a hypothesis statement. Laurer et al. \cite{laurer2024} fine-tune a DeBERTa model on multiple NLI datasets, producing a general-purpose classifier that can be applied to any classification task by expressing each candidate label as a natural language hypothesis. For stance detection, the input message serves as the premise and each candidate stance label, "this text expresses a pro-Palestine stance," "this text expresses a pro-Israel stance," "this text is neutral", serves as a hypothesis. The label with the highest entailment score is selected. This requires no labeled stance data and generalizes across tasks, but its performance depends on how closely the vocabulary and framing of the input aligns with what the NLI model learned during pre-training.

\subsubsection{Fine-Tuned Stance Detection with BERTweet}
\label{sec:stance-bertweet}
The fine-tuned stance detection model uses BERTweet \cite{nguyen2020} as its base. BERTweet shares BERT-base's architecture but is pre-trained from scratch on roughly 850 million English tweets using the RoBERTa procedure, with tweet-specific preprocessing (normalising user mentions and URLs, converting emoji to text) and a vocabulary built specifically from tweet text, so that informal language, slang, and hashtags are tokenised with fewer out-of-vocabulary splits than a vocabulary built from formal text \cite{nguyen2020}. Nguyen et al. show it outperforms RoBERTa-base on tweet-specific tasks such as sentiment analysis and irony detection, confirming that domain-specific pre-training provides a measurable advantage over larger general-purpose models \cite{nguyen2020}. For stance classification in this study, BERTweet's pre-trained transformer layers are retained and a three-class linear classification layer is appended. During fine-tuning, the [CLS] token's final hidden state, a 768-dimensional vector encoding the full contextual representation of the input sequence, is passed through the linear layer and trained with cross-entropy loss on the 736 manually labeled messages described in Section~\ref{sec:dataset}. All transformer weights are updated during fine-tuning, not just the classification layer, allowing the domain-specific representations built during tweet pre-training to be further adapted to the political framing patterns specific to this dataset.

\subsubsection{Political Stance Analysis and Dataset Challenges}
\label{sec:stance-challenges}
Political stance in conflict contexts differs from the standard benchmarks on which most methods are evaluated. SemEval-2016 targets involve abstract social and political topics, where stance is often expressed through opinion directed at a named target \cite{mohammad2016}. Conflict-related Telegram content differs in at least three ways. First, the dominant register is reportorial: stance is embedded in vocabulary selection and actor framing rather than in explicit political declarations. Second, multi-actor references, to Iranian forces, Hezbollah, the UN, Western governments, signal stance through association rather than explicit endorsement. Third, the annotation boundary between neither and politically aligned content is unstable in a corpus where all channels are politically committed, since messages that appear superficially neutral often contain implicit framing that annotators assign inconsistently. This instability at the neutral boundary is a recognised difficulty in the stance annotation literature \cite{kucuk2020}, and it anticipates the weak neutral-class performance reported across all three methods in Section~\ref{sec:evaluation}.

\subsection{Narrative Analysis in Online Communities}
\label{sec:narrative}

\subsubsection{From Strategic Narrative to Community-Level Pattern}
\label{sec:narrative-strategic}
Section~\ref{sec:infowar} established that conflict actors construct strategic narratives to legitimize their position and delegitimize opponents, operating simultaneously at the system, national, and issue levels \cite{roselle2014}. That analysis focused on deliberate, institutionally coordinated communication, state ministries, military spokespersons, organized media operations. The Telegram channels in this dataset are something different: communities of politically aligned administrators and subscribers who sustain narrative structures not through coordinated strategy but through the accumulated effect of consistent framing choices repeated across thousands of individual messages over years. The distinction matters analytically. A single message expressing solidarity with Palestinian civilians, or describing an Israeli military operation in operational terms, is not in itself a narrative. It becomes part of one through repetition: when the same vocabulary, the same actor categories, the same moral attributions appear across hundreds of messages from the same channel, they constitute a collectively maintained interpretive schema through which subscribers encounter new events. As Lewandowsky et al. note, such frames are cognitive tools for managing complexity \cite{lewandowsky2013}; in the Telegram channel format, administrators perform this reduction on behalf of subscribers, selecting which events to report and which framing attributes to foreground.

\subsubsection{Framing as Measurable Attribute Selection}
\label{sec:narrative-framing}
Entman's definition of framing as the selection and salience of aspects of perceived reality, promoting a particular problem definition, causal interpretation, moral evaluation, or treatment recommendation \cite{entman1993}, provides the operational vocabulary for this study's framing analysis. The four framing categories constructed in Section~\ref{sec:methodology} (death context, victim framing, military framing, legal framing) each correspond to one of Entman's functional dimensions. They are not arbitrary topic labels but measures of which attributes channels systematically foreground when describing the same events. Scheufele draws on the distinction between agenda-setting and frame-setting: where agenda-setting concerns the salience of issues, frame-setting, which McCombs and colleagues equate with second-level agenda-setting, concerns the salience of issue attributes, the specific dimensions through which an issue is interpreted \cite{scheufele1999}. Pro-Israel and pro-Palestine channels cover the same military operations, the same casualty events, the same legal proceedings, the divergence is not in topic coverage but in attribute selection. Tracking which framing categories dominate each channel group, and correlating them with sentiment distributions, provides an empirical measure of how the same events are positioned differently within each community's narrative structure.

\subsubsection{Alternative Media Positioning and Identity}
\label{sec:narrative-altmedia}
The channels in this dataset explicitly position themselves against mainstream media coverage, a structural feature that Haller, Holt, and de la Brosse identify as definitional to alternative media: outlets run by actors who publish interpretations of current events as a direct response to the perception that their perspective is being treated unfairly in the mainstream media \cite{haller2019}. This oppositional self-positioning reinforces the identity function of the channels. Subscribers are not simply receiving information, they are addressed as members of a community whose perspective is systematically suppressed elsewhere, and the channel's consistent framing choices constitute and reinforce that collective identity over time. This is why the framing analysis in Section~\ref{sec:methodology} is not conducted on individual messages but at the aggregate level across stance groups. Individual messages are data points; the narrative structure is the distributional pattern. A channel that consistently applies legal framing, invoking international law, the ICC, war crimes, is not just reporting on legal developments in specific messages. It is constructing a community-level narrative in which the conflict is fundamentally a question of accountability and criminality. A channel that dominates in military framing with neutral sentiment is constructing a different narrative: one in which military operations are procedural facts requiring documentation rather than moral events requiring evaluation. These patterns only become visible at scale, which is precisely what the 87,617-message corpus makes possible. The theoretical and computational frameworks presented in this section, agenda-setting and framing theory, strategic narrative, the architecture of Telegram as broadcast infrastructure, and the sentiment-analysis and stance-detection methods on which this study relies, together form the conceptual and methodological foundation of the study. The following chapter situates the present work within existing empirical research, comparing its approach, data, and findings against prior studies of conflict communication, before Section~\ref{sec:dataset} describes the dataset and Section~\ref{sec:methodology} details how these computational methods were implemented and adapted to the analysis of Israel--Palestine Telegram discourse.

\section{Related Work}
\label{sec:related}
This section situates the present study within existing research. Where Section~\ref{sec:background} introduced the theoretical and computational concepts the study relies on, this section surveys prior empirical work, comparing the approaches, data, and findings of earlier studies and positioning the present analysis against them. The review is organized into five thematic areas, Telegram as a conflict communication platform, social media studies of the Israel--Palestine conflict, sentiment analysis in political communication, stance detection, and narrative analysis, followed by a synthesis that states the gap this study addresses.

\subsection{Telegram Conflict Studies}
\label{sec:related-telegram}
Research on Telegram as a venue for conflict-related communication has grown alongside the platform's adoption by politically motivated actors. Early work focused on its role as moderation-resistant infrastructure: Prucha \cite{prucha2016} documented the migration of Islamic State media operations from Twitter to Telegram as Twitter's content moderation intensified, arguing that Telegram's encryption, large-file sharing, and resistance to deletion made it structurally attractive to actors facing suppression elsewhere. Dargahi Nobari et al. \cite{dargahi2017} approached the platform from a different angle, providing a large-scale structural analysis of Telegram's channel architecture and growth, and establishing the broadcast, non-social-graph characteristics that distinguish it from algorithmically curated networks. More recent work has examined Telegram specifically in the context of the Israel--Palestine conflict. Antonakaki and Ioannidis \cite{antonakaki2025telegram} conducted a cross-platform analysis spanning Telegram, Reddit, and Twitter following October 7, 2023, identifying distinct discourse communities with different sentiment profiles and documenting a measurable increase in Telegram use among both Israeli and Palestinian audiences. A subsequent study by the same authors \cite{antonakaki2026crossplatform} characterized Telegram as a hub for uncensored, real-time documentation during the conflict. These studies establish Telegram as a significant conflict-communication platform, but they share a common feature: they treat Telegram as one node among several in a cross-platform comparison, rather than as the dedicated object of study. Two features distinguish the present work from these studies. It treats Telegram channels as the dedicated object of analysis rather than one node in a cross-platform comparison, enabling a deeper view of channel-level discourse; and it draws on a five-year corpus (May 2021--June 2026) spanning multiple escalations, where prior Telegram studies concentrate on the weeks following October 2023.

\subsection{Israel--Palestine Social Media Studies}
\label{sec:related-ip}
A substantial body of work has examined how the Israel--Palestine conflict is communicated and contested on social media, predating the computational turn this study represents. Siapera, Hunt, and Lynn \cite{siapera2015} analysed roughly three million tweets from the 2014 Gaza war (Operation Protective Edge), characterizing the discourse through the concepts of mediality and vectors and finding it dominated by activist and witness accounts, a pronounced pro-Palestinian volume, and a humanitarian register cutting across both partisan poles. Manor and Crilley \cite{manor2018} approached the same 2014 conflict from the state-actor side, analysing the Israeli Ministry of Foreign Affairs' Twitter activity and identifying fourteen distinct frames deployed to legitimize Israeli policy while delegitimizing Hamas. Zeitzoff \cite{zeitzoff2017} examined the earlier 2012 Gaza conflict, documenting how both Israel and Hamas ran parallel information campaigns on Twitter alongside military operations. More recently, Osimen et al. \cite{osimen2025} analysed asymmetric media framing in the aftermath of October 7, 2023, finding that Western outlets predominantly adopted Israeli narratives while Palestinian perspectives received comparatively limited coverage. Two patterns are visible across this work. First, it is concentrated overwhelmingly on Twitter and on single conflict episodes, the 2012 and 2014 Gaza wars, or the immediate aftermath of October 2023. Second, the dominant methods are qualitative or descriptive (frame analysis, mediality/vector analysis, content analysis) rather than large-scale computational classification. Where sentiment is measured, as in Siapera et al. \cite{siapera2015}, it is reported at the aggregate level without stance- or framing-conditioned breakdowns. Against this largely qualitative, single-episode literature, the present work contributes scale and method: multi-method computational classification (three stance methods, sentiment, and framing) over a multi-escalation corpus, which yields the sentiment-within-framing breakdowns descriptive studies could not produce. Its central result, that the two sides deploy the same death- and victim-framing vocabulary in opposite emotional registers (Section~\ref{sec:discussion}), turns Siapera et al.'s \cite{siapera2015} observation of a cross-cutting humanitarian register into a quantified, stance-conditioned finding.

\subsection{Sentiment Analysis in Political Communication}
\label{sec:related-sentiment}
Sentiment analysis has been widely applied to political communication, including electoral discourse, protest movements, and conflict reporting. The methodological foundations span lexicon-based approaches \cite{taboada2011}, classical supervised machine learning \cite{pang2002}, and, more recently, transformer-based models that capture context and compositional meaning beyond the reach of bag-of-words representations \cite{socher2013}, \cite{vaswani2017}. Transformer-based sentiment models have been applied across the spectrum of political communication, from electoral discourse to active armed conflict. In the electoral domain, Khan et al. \cite{khan2023elecbert} fine-tuned a BERT-based model (ElecBERT) on millions of US election tweets and, in a case study of the 2020 presidential election, found that political tweets were predominantly classified as neutral, attributing this to the prevalence of objective, news-style statements over direct emotional expression. Sentiment analysis of conflict-related discourse shows a similar reliance on transformer architectures while exhibiting a markedly more negative emotional profile. Ramos and Chang \cite{ramos2023} applied a RoBERTa-based model to roughly 600,000 English tweets concerning the Russia--Ukraine war, finding fear (32\%) and anger (15\%) to be the dominant emotional categories, with vocabulary centred on casualties, weapons, and crisis. Hasan et al. \cite{hasan2023} analysed Bangla-language social-media comments on the same conflict using several transformer models, with a fine-tuned in-domain model (BanglaBERT) achieving the strongest performance (macro-F1=0.82); their distribution was again neutral-dominant, and their weakest class was a politically aligned one whose under-performance they attribute to class imbalance and implicit, indirectly expressed sentiment. Two patterns recur across these applications and are directly relevant to the present study. First, transformer models pre-trained on or fine-tuned for the target domain consistently outperform general-purpose alternatives, mirroring the stance-detection findings of Section~\ref{sec:related-stance}. Second, aggregate sentiment in political and conflict corpora is dominated by neutral content, even in partisan or crisis-driven sources, because much of the material consists of event reporting rather than overt emotional expression, and the classes carrying implicit rather than explicit sentiment are the hardest to classify. Both patterns anticipate results reported later in this study: the predominance of neutral sentiment in the corpus (Section~\ref{sec:results}) and the comparative difficulty of the neutral stance class across all three detection methods. Here too the present work adopts a Twitter-pretrained sentiment model \cite{barbieri2020} suited to the informal register of Telegram messages, but rather than treating sentiment as the endpoint, it conditions sentiment on model-assigned stance and framing category, surfacing register differences (Sections~\ref{sec:results}--\ref{sec:discussion}) that aggregate sentiment alone conceals.

\subsection{Stance Detection}
\label{sec:related-stance}
Stance detection, determining the position a text takes toward a specified target, has evolved from feature-engineered classifiers to transformer-based models. The task was formalized at scale by the SemEval-2016 shared task \cite{mohammad2016}, whose SVM baseline outperformed all participating systems, and surveyed comprehensively by Küçük and Can \cite{kucuk2020}, who document the dominance of feature-based SVM approaches in early work and the subsequent shift toward deep learning. With the introduction of transformer-based models, stance detection shifted decisively toward fine-tuned pre-trained language models. Li et al. \cite{li2021pstance} introduced P-Stance, a large political stance dataset of 21,574 tweets targeting Donald Trump, Joe Biden, and Bernie Sanders, and established a strong fine-tuned baseline: BERTweet, a model pre-trained specifically on tweets, achieved a macro-averaged F1 of 80.5\%, outperforming a fine-tuned BERT (76.5\%) and all non-transformer baselines such as BiLSTM and CNN architectures. The advantage of in-domain pre-training is corroborated by Kawintiranon and Singh \cite{kawintiranon2022}, who compared RoBERTa, TweetEval, and BERTweet on 2020 US election stance data and found the tweet-pre-trained models consistently stronger than the general-purpose RoBERTa, with further gains from domain-specific continued pre-training. The dominance of fine-tuned transformers, and of tweet-pre-trained models in particular, is visible most clearly in competitive evaluations. In the CASE 2024 shared task on stance detection in climate-activism tweets \cite{case2024}, where nineteen teams applied a range of architectures to a common dataset, the top-performing system was a fine-tuned BERTweet model (macro-F1 74.8\%), narrowly ahead of a DeBERTa-based system (73.9\%); transformer-based encoders dominated the leaderboard overall. Across these studies a consistent ordering emerges: fine-tuned in-domain transformers outperform general-purpose transformers, which in turn outperform lexicon- and feature-based methods. Notably, this advantage largely persists even against much larger models. Garg and Caragea \cite{garg2024}, evaluating BERT-based models against autoregressive large language models including Llama-2 and GPT-3.5 across four stance datasets, found that in full-dataset fine-tuned settings the large language models generally did not surpass fine-tuned BERT-based models, although GPT-3.5 proved competitive and outperformed them on one zero-shot benchmark. This suggests that for stance detection on short, informal text with available labeled data, in-domain fine-tuning of a comparatively small model remains a strong and computationally efficient choice. Across this literature a consistent finding emerges, summarized by Küçük and Can \cite{kucuk2020}: fine-tuned, in-domain transformer models outperform both lexicon-based and zero-shot approaches when even modest amounts of labeled data are available. This finding directly motivates the methodology of the present study. These findings motivate the use of BERTweet here, and the study adds to this line of work in two ways: it runs a direct, controlled comparison of the three stance paradigms on a single corpus against shared ground truth, isolating method from data; and it applies them to a domain, pro-Israel/pro-Palestine Telegram, with its reportorial register and multi-actor regional framing, where the gap between in-domain fine-tuning and the label-free approaches proves especially pronounced (Section~\ref{sec:results}).

\subsection{Narrative Analysis in Conflict Communication}
\label{sec:related-narrative}
A parallel research tradition approaches conflict communication through narrative and framing rather than classification. Roselle, Miskimmon, and O'Loughlin \cite{roselle2014} developed strategic narrative theory, arguing that combatants construct meaning at the system, national, and issue levels, and that digitization destabilizes the settled meaning of past events. Lewandowsky et al. \cite{lewandowsky2013} examined the cognitive function of conflict narratives, showing both that they serve as necessary tools for managing complexity and that misinformation embedded in a dominant narrative can persist for years after correction. On the structural side, Haller, Holt, and de la Brosse \cite{haller2019} characterized alternative media as outlets that position themselves correctively against perceived mainstream bias, a stance the channels in this dataset adopt directly. This tradition is theoretically rich but methodologically distinct from the present work: it is largely interpretive, analysing narratives through close reading rather than measuring their distribution at scale. Manor and Crilley's \cite{manor2018} frame analysis, for instance, identifies fourteen frames through qualitative coding of a bounded set of messages. The strength of this approach is depth; its limitation is that narrative claims are difficult to verify across large corpora. The present work operationalizes these narrative-analytic concepts computationally: rather than identifying frames through close reading, it defines four framing categories as measurable indicators and tracks their distribution across the corpus, conditioned on model-assigned stance. This trades interpretive depth for scale and statistical testability, the two approaches being complementary, the qualitative tradition explaining what frames mean and the computational one measuring how often, and by whom, they are deployed.

\subsection{Summary and Research Gap}
\label{sec:related-gap}
The studies reviewed in this section establish three things. First, Telegram is now a significant platform for conflict-related political communication, but existing Telegram studies of the Israel--Palestine conflict treat it as one node in cross-platform comparisons rather than as a dedicated object of study, and concentrate on the period immediately after October 2023 \cite{antonakaki2025telegram}, \cite{antonakaki2026crossplatform}. Second, social media analysis of the conflict is overwhelmingly concentrated on Twitter and on single conflict episodes, using predominantly qualitative or descriptive methods \cite{zeitzoff2017}, \cite{osimen2025}, \cite{manor2018}, \cite{siapera2015}. Third, while transformer-based methods now dominate both sentiment analysis and stance detection, they have not been systematically applied or compared on a dedicated pro-Israel/pro-Palestine Telegram corpus. The gap this study addresses follows directly. No published study has conducted a dedicated, multi-method computational analysis comparing pro-Israel and pro-Palestine Telegram channels over an extended period spanning multiple escalations. It does so with the corpus and methods set out above, providing what is, to our knowledge, the first dedicated comparison of this kind and extending the descriptive findings of prior Twitter-based studies into quantified, stance-conditioned results.

\section{Dataset}
\label{sec:dataset}

\subsection{Data Source}
\label{sec:data-source}
Data for this study was collected exclusively from Telegram. The platform has played a major role in conflict-related information sharing over the past decade, including the Syrian civil war and the campaign of SAA against ISIS, the war in Ukraine, and the Israel--Palestine conflict. Telegram channels are one-directional broadcasting mechanisms: administrators publish content and subscribers receive it in chronological order, without algorithmic curation or engagement-based ranking. Channel content therefore mirrors the administrators' own posting decisions, not the output of a recommendation system.

\subsection{Scope: The Axis of Resistance as Context}
\label{sec:scope}
The dataset was not restricted to channels discussing Gaza and the West Bank exclusively. The Palestinian struggle is frequently framed, by both supporters and opponents, in relation to a broader regional alliance, Iran together with Hezbollah, the Houthi movement in Yemen, and aligned factions in Iraq, collectively known as the "axis of resistance" \cite{mansour2025}. This framing is not merely an analytical abstraction; it appears directly and consistently in the Telegram content examined. Channels aligned with these actors routinely discussed developments in Gaza alongside their own military operations, presenting them as fronts within a single, unified conflict rather than as separate regional crises. This dynamic became especially pronounced following the Israeli strike on the Iranian consulate in Damascus in April 2024 and the Iranian missile exchanges that followed. Both pro-Palestine and pro-Israel channels treated these events as continuous with the October 2023 escalation, not as a distinct interstate confrontation. Because all actors within the axis of resistance publicly ground their operations in the Palestinian cause, their channels constitute a direct source of pro-Palestine narrative and sentiment. Excluding them would have produced a cleaner dataset in terms of geographic scope, but a substantially less representative one in terms of how the conflict is actually communicated and framed by its participants.

\subsection{Channel Selection}
\label{sec:channels}
Sixteen Telegram channels were selected manually: 8 pro-Palestine and 8 pro-Israel. Selection was based primarily on subscriber count, posting frequency, and clear ideological alignment. These channels had been monitored over several years prior to the study, which provided familiarity with their political positioning and content. Channels emphasizing civilian casualties, occupation, resistance, and international law violations were categorized as pro-Palestine. Channels emphasizing Israeli security operations, military necessity, and counter-terrorism were categorized as pro-Israel. The obvious limitation is that channel selection relies on the author's judgment. An automated channel-discovery process would still require manual validation at the labeling stage. The advantage of manual selection is that it ensures the inclusion of influential and highly active channels rather than obscure accounts with limited reach.

\subsection{Data Collection}
\label{sec:data-collection}
Messages were collected using a custom Python crawler provided by the study supervisor. The crawler retrieved posts in reverse chronological order, beginning with the most recent messages and continuing until approximately 47,000 messages had been collected from each stance group. This even split by source was deliberate: collecting comparable volumes from each side avoids building in a structural imbalance that could bias the aggregate sentiment, stance, and framing comparisons toward whichever group was more heavily sampled. After removing empty and non-text entries, the initial corpus contained 94,290 messages. Each record included the message text, timestamp, post identifier, channel name, and stance-group label. The dataset spans May 2021 to June 2026, although most messages were published after October 2023 because collection began with the newest available posts.

\subsection{Preprocessing}
\label{sec:preprocessing}
Two preprocessing steps were applied before analysis: Deduplication Reposting is common on Telegram, particularly during periods of intense activity. Messages were normalized by converting text to lowercase, removing URLs, and stripping punctuation. Exact duplicates were then identified and removed while retaining the first occurrence. This process removed 5,712 messages (6.1\%), reducing the dataset from 94,290 to 88,578 messages. The relatively low duplication rate suggests that although channels often shared similar narratives, they were not functioning primarily as reposting networks. Language Filtering Because the NLP models used in this study were designed for English-language text, non-English messages were removed. Language identification was performed using the langdetect library \cite{danilak2014}. A total of 961 messages (1.1\%) were excluded, resulting in a final dataset of 87,617 messages.

\subsection{Dataset Composition}
\label{sec:composition}
Each record in the final dataset contains the message text, timestamp, channel name, platform identifier, post identifier, and channel label. Sentiment and stance-related variables were added later in the analysis pipeline and are described in Section~\ref{sec:methodology}.

\begin{table}[t]
\centering
\caption{Preprocessing pipeline.}
\label{tab:preprocessing}
\begin{tabular}{lrr}
\toprule
Stage & Messages & Removed \\
\midrule
After collection & 94{,}290 & -- \\
After deduplication & 88{,}578 & 5{,}712 (6.1\%) \\
After language filtering & 87{,}617 & 961 (1.1\%) \\
Final dataset & 87{,}617 & -- \\
\bottomrule
\end{tabular}
\end{table}

\subsection{Ethical Considerations}
\label{sec:ethics}
All data used in this study was obtained from publicly accessible Telegram channels. No private conversations, restricted groups, or user-level data were collected. No personally identifiable information was stored or analyzed. Throughout the study, the unit of analysis is the channel rather than individual users.

\subsection{Manual Annotation}
\label{sec:annotation}
A manually labeled dataset was created to train and evaluate the supervised stance-classification model. The initial target was 1,000 annotated messages. In practice, the final sample contained 736 messages. Labeling those messages individually, without a second annotator to cross-check decisions, took considerably longer than the initial estimate. Each message was read and assigned one of three labels: pro-Palestine, pro-Israel, or neutral. The most difficult cases involved channels quoting opposing viewpoints, posts combining factual reporting with implicit framing, and messages where stance was conveyed indirectly rather than through explicit language. Following the initial annotation pass, the dataset contained an over-representation of pro-Israel examples. To reduce class imbalance, samples from the majority class were removed. The final labeled dataset consisted of 736 messages: 253 pro-Palestine, 253 pro-Israel, and 230 neutral. The slightly smaller neutral class reflects the nature of the source material. Because all messages originated from politically aligned channels, genuinely neutral content was comparatively uncommon. This labeled dataset was used for two purposes: training the fine-tuned BERTweet model and evaluating the performance of all stance-detection approaches against human-assigned labels. The cross-validation and evaluation procedure are described in Section~\ref{sec:methodology}.

\section{Methodology}
\label{sec:methodology}

\subsection{Overview}
\label{sec:method-overview}
\begin{figure}[t]
\centering
\includegraphics[width=0.85\linewidth]{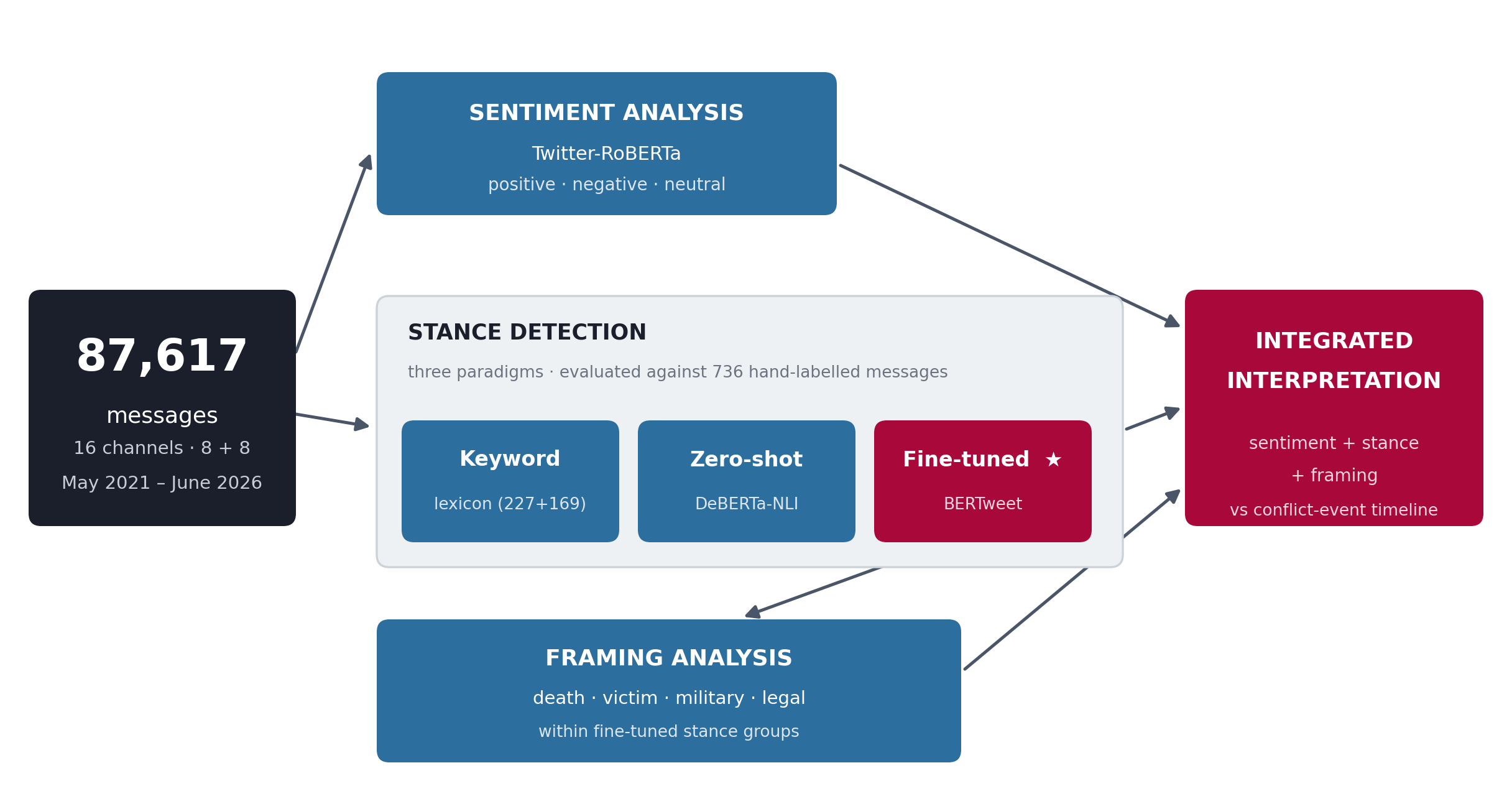}
\caption{Analysis pipeline.}
\label{fig:5_1}
\end{figure}

The analytical pipeline consists of four main components: sentiment analysis, keyword-based stance detection, zero-shot stance detection, and fine-tuned stance detection. These methods were applied to the 87,617-message dataset described in Section~\ref{sec:dataset}. A framing analysis was also conducted to examine how different stance groups discuss recurring themes such as civilian casualties, military operations, and legal accountability. The three stance detection approaches were chosen to represent the three principal paradigms in the evolution of stance detection, as surveyed in Section~\ref{sec:related-stance}: lexicon- and feature-based classification (the keyword method), zero-shot transfer via natural language inference (the DeBERTa model), and fine-tuned in-domain transformers (the BERTweet model). Comparing one representative of each paradigm on the same corpus and against the same ground truth serves two purposes. First, it isolates the effect of method while holding the data constant, allowing a direct assessment of how much in-domain fine-tuning gains over approaches that require no labeled data. Second, the lexicon and zero-shot methods, which need no training data, function as informative baselines: where they succeed, stance is carried by explicit surface signals; where they fail but fine-tuning succeeds, stance is carried by implicit or domain-specific framing. The methods were designed to be directly comparable, with each assigning one of three labels, pro-Palestine, pro-Israel, or neutral, to each message. Their predictions were evaluated against the 736 manually annotated messages described in Section~\ref{sec:annotation}. This allows the methods to be compared both in terms of performance and in the types of classifications they produce. Figure~\ref{fig:5_1} summarizes the overall pipeline structure. The complete implementation of the pipeline, preprocessing, the three stance detectors, the sentiment and framing analyses, and the scripts that produce the reported tables and figures, is publicly available \cite{zafeiropoulos2026}.

\subsection{Sentiment Analysis}
\label{sec:sentiment}

Sentiment analysis was applied to the full dataset before stance detection. Each message was classified as positive, negative, or neutral using the cardiffnlp/twitter-roberta-base-sentiment-latest model through the Hugging Face Transformers library. The model is based on RoBERTa and was fine-tuned on the TweetEval benchmark. TweetEval \cite{barbieri2020} provides a unified benchmark for several Twitter-based classification tasks, including sentiment analysis. This model was selected because it aligns with the data on two axes simultaneously: it is pre-trained on a large tweet corpus, matching the informal social-media register of the Telegram messages, and it is fine-tuned specifically for sentiment classification, matching the task. It is also the established, maintained sentiment model derived from the TweetEval benchmark \cite{barbieri2020}, making it a standard choice for tweet-style sentiment analysis, and its tweet-based pre-training keeps the sentiment component consistent with the tweet-pre-trained foundation used for stance detection (Section~\ref{sec:finetuned}). The choice of a social-media-trained model was based on the characteristics of the dataset. Telegram messages from politically active channels share several similarities with tweets: they are generally short, informal, and frequently contain hashtags, named entities, abbreviations, and emotionally charged language. Models trained primarily on formal text sources such as news articles or reviews are less suited to this type of content. This pre-trained model was used directly, without task-specific fine-tuning, in contrast to the stance detection component (Section~\ref{sec:finetuned}), where a model was fine-tuned on manually annotated data. This asymmetry was a deliberate methodological choice. Sentiment polarity is a general, well-resourced classification task for which robust pre-trained models trained on large tweet corpora are readily available, and for which a custom model would offer little benefit. Stance toward the Israel--Palestine conflict, by contrast, is a domain-specific task with no suitable pre-trained model available, making fine-tuning on in-domain annotated data necessary. Annotation effort was therefore concentrated on the task where it was most needed and where it could be expected to yield the greatest improvement over off-the-shelf alternatives. Messages were truncated to a maximum of 512 tokens where required, as this represents the maximum input length supported by the model. In practice, the vast majority of messages fell well within this limit. These outputs were stored and used in later analysis.

\subsection{Keyword-Based Stance Detection}
\label{sec:keyword}

The first stance detection approach uses a weighted keyword matching method. Two manually constructed lexicons were created: one representing pro-Palestine framing and one representing pro-Israel framing. Each message was scored against both lexicons, and the final stance label was determined from the resulting scores.

\subsubsection{Lexicon Construction}
\label{sec:lexicon}
The lexicons were developed through direct observation of the Telegram channels included in the dataset. The author has followed many of these channels over several years across multiple conflicts, providing familiarity with recurring terminology, narratives, and framing patterns used by each side. This type of domain-specific lexicon construction is commonly used in political text analysis, where certain terms carry ideological meaning that may not be captured by general-purpose language models. The lexicons were also informed by patterns observed in news reporting, opinion pieces, and broader public discourse surrounding the conflict. Some terms appeared almost exclusively in one political context, while others appeared across both groups but carried different meanings depending on usage. Each keyword was assigned a weight based on how strongly it indicated stance. Weight 3 was assigned to highly distinctive terms that rarely appeared outside one side's discourse. Examples from the pro-Palestine lexicon include "free palestine", "ethnic cleansing", "genocide", "open air prison", and "from the river to the sea". Examples from the pro-Israel lexicon include "october 7 massacre", "never again", "am yisrael chai", and "hamas is isis". Weight 2 was assigned to strong indicators that could occasionally appear in opposing or neutral contexts. Examples include "apartheid", "war crimes", "nakba", "occupation forces", and "settler violence" on the pro-Palestine side, and "antisemitism", "human shields", "iron dome", "hostages", and "terror tunnels" on the pro-Israel side. Weight 1 was assigned to weaker indicators that appeared in broader discussion but still leaned toward one stance. Examples include "ceasefire", "gaza", "displaced", and "refugee camp" for the pro-Palestine lexicon, and "idf operation", "precision strike", and "two-state solution" for the pro-Israel lexicon. The final pro-Palestine lexicon contained 227 terms, while the pro-Israel lexicon contained 169 terms. The classification follows a lexicon-based scoring approach, in which stance is inferred by comparing the cumulative weight of matched terms from each lexicon.

\subsubsection{Scoring and Classification}
\label{sec:scoring}
Before classification, message text was normalized by converting it to lowercase, removing URLs, and collapsing unnecessary whitespace. Term matching was applied on word boundaries, with longest-match precedence when one listed term was contained within another. This prevents substring false positives (for example, "infant" matching inside "infantry") and ensures that multi-word terms drawing vocabulary from both lexicons (for example, "houthi missiles") resolve to a single side instead of cancelling to neutral. Each message was scored by identifying matching terms from both lexicons and summing their corresponding weights. Messages with a higher pro-Palestine score were classified as pro-Palestine, while messages with a higher pro-Israel score were classified as pro-Israel. Messages where both scores were equal, or where no matching terms were found, were classified as neutral.

\subsection{Zero-Shot Stance Detection}
\label{sec:zeroshot}
The second stance detection approach uses zero-shot classification with the MoritzLaurer/deberta-v3-large-zeroshot-v2.0 model. Zero-shot classification allows a model to assign labels without being trained on task-specific examples. Instead, the model evaluates how well a text matches a set of candidate labels using natural language inference. The specific checkpoint (deberta-v3-large-zeroshot-v2.0) is a DeBERTa-v3-large model fine-tuned for natural language inference following the NLI-based zero-shot approach of Laurer et al. \cite{laurer2024}, which makes it suitable for zero-shot classification without task-specific training. This model was selected for two reasons. First, the zero-shot method adopted here is itself NLI-based, requiring a model fine-tuned for natural language inference; DeBERTa-v3-large-NLI is the standard and one of the strongest publicly available checkpoints for this purpose, and is the architecture around which the approach of Laurer et al. \cite{laurer2024} is built. Because zero-shot classification accuracy depends directly on the quality of the underlying entailment model, using a leading NLI model is the principled choice, and DeBERTa's documented advantage over RoBERTa on natural language inference (Section~\ref{sec:stance-deberta}) makes it well suited to the role. Second, using a strong general-purpose model contrasts deliberately with the tweet-specific fine-tuned model, isolating the effect of in-domain adaptation rather than of architecture alone. The model is based on DeBERTa (Decoding-enhanced BERT with Disentangled Attention) \cite{deberta2021}. DeBERTa builds on BERT \cite{devlin2019} and RoBERTa \cite{liu2019}, modifying the attention mechanism to better represent content and positional information. For each message, the model was given three possible labels: pro\_palestine, pro\_israel, and neutral. The label with the highest confidence score was assigned as the predicted stance. Multi-label classification was disabled, meaning each message received a single final classification. This method was included as a baseline that does not rely on the manually annotated Telegram data. A limitation is that the model has not been adapted to the vocabulary, terminology, and framing patterns that appear frequently in the dataset.

\subsection{Fine-Tuned Stance Detection}
\label{sec:finetuned}
The third stance detection method fine-tunes a pre-trained language model using the manually annotated dataset described in Section~\ref{sec:annotation}. The base model selected was BERTweet \cite{nguyen2020}.

\subsubsection{Model Selection}
\label{sec:model-selection}
BERTweet is a transformer-based language model built on the BERT architecture introduced by Devlin et al. \cite{devlin2019}. It was pre-trained on approximately 850 million English tweets and was designed specifically for social media text \cite{nguyen2020}. The model was selected because Telegram messages share many characteristics with Twitter data. Messages in the dataset are often short, informal, and contain hashtags, abbreviations, named entities, and conversational language. A model pre-trained on similar text provides a more suitable starting point than models trained mainly on formal documents.

\subsubsection{Training Setup}
\label{sec:training}
The 736 manually labeled messages were divided using 5-fold stratified cross-validation. Stratification ensured that the class proportions of pro-Palestine, pro-Israel, and neutral examples were preserved across all folds. In each fold, 80\% of the labeled data (\textasciitilde{}589 examples) was used for training and the remaining 20\% (\textasciitilde{}147 examples) served as the held-out validation set. Each fold trained an independent model initialized from BERTweet-base with no weight transfer between folds. Text was tokenized using the BERTweet tokenizer with a maximum sequence length of 128 tokens. Shorter sequences were padded and longer sequences were truncated. Each fold was trained for three epochs with a batch size of 16. Performance was reported as the mean and standard deviation of macro F1 and accuracy across the five folds. To produce leakage-free predictions for the labeled rows, out-of-fold (OOF) predictions were collected: each labeled message received a prediction from the fold in which it appeared in the held-out 20\%, ensuring it was never seen during that model's training. A final model was then trained on all 736 labeled examples and applied to the complete dataset of 87,617 messages. For the labeled rows, the final model's predictions were replaced with their corresponding OOF predictions to avoid in-sample optimism in the evaluation.

\subsubsection{Computational Environment}
\label{sec:compute}
All experiments were conducted on a personal Apple Silicon MacBook (M4). Models supporting Apple's Metal Performance Shaders (MPS) backend were run with GPU acceleration where available. The zero-shot classification model was executed on CPU due to compatibility limitations and required approximately three days to process the full dataset. Fine-tuning BERTweet completed within a few hours using MPS acceleration.

\subsubsection{Label Scheme}
\label{sec:label-scheme}
The same three-class scheme used throughout this study was applied during fine-tuning: pro-Palestine, pro-Israel, and neutral. The original output layer of BERTweet was replaced with a three-class classification layer initialized for this task. The transformer layers retained their pre-trained weights and were updated during training.

\subsection{Framing Analysis}
\label{sec:framing-method}

Framing analysis was conducted on the full dataset to examine how different stance groups discuss specific aspects of the conflict. Four framing categories were defined:

\begin{itemize}
    \item \textbf{Death context}: Captures references to casualties, killings, and physical destruction, including terms such as \emph{``killed''}, \emph{``massacre''}, \emph{``death toll''}, and \emph{``airstrike''}.

    \item \textbf{Victim framing}: Focuses on language emphasizing civilian suffering and vulnerability, including terms such as \emph{``children''}, \emph{``hospital''}, \emph{``refugees''}, \emph{``displaced''}, and \emph{``aid workers''}.

    \item \textbf{Military framing}: Captures operational language related to military activity, including terms such as \emph{``precision strike''}, \emph{``ground offensive''}, \emph{``combat mission''}, and military organization names.

    \item \textbf{Legal framing}: Includes references to international law and accountability mechanisms, including terms such as \emph{``war crimes''}, \emph{``ICC''}, \emph{``ICJ''}, \emph{``Geneva Convention''}, and \emph{``arrest warrant''}.
\end{itemize}

Each message was matched against these framing categories using keyword-based classification. Because framing dimensions are not mutually exclusive, a single message could receive multiple framing labels. The resulting framing assignments were subsequently analyzed together with the sentiment and stance classifications to examine how different communities framed the conflict over time.

\subsection{Evaluation}
\label{sec:evaluation-method}
All three stance detection methods were evaluated against the 736 manually annotated messages described in Section~\ref{sec:annotation}. Two evaluation metrics were used: accuracy and macro F1 score. Accuracy measures the proportion of correctly classified messages across all classes. Macro F1 calculates the average F1 score across the three classes while giving equal importance to each category. Macro F1 was selected because the annotated dataset contains slightly different numbers of examples per class, and the metric provides a clearer indication of whether a method performs consistently across all categories. Confusion matrices were generated for each model to identify common error patterns, particularly whether errors occurred between the two opposing stance classes or whether neutral messages were incorrectly assigned politically aligned labels. Finally, inter-method agreement was calculated across the complete dataset. For each pair of methods, the percentage of messages receiving identical labels was measured. This shows whether the methods produce similar labels when applied to the full dataset.

\section{Results}
\label{sec:results}

\subsection{Sentiment Analysis}
\label{sec:results-sentiment}
Sentiment analysis was applied to all 87,617 messages using the Twitter-RoBERTa model described in Section~\ref{sec:sentiment}. The overall distribution was dominated by neutral sentiment, which accounted for approximately 64\% of messages. Negative sentiment represented around 32\% of the dataset, while positive sentiment was uncommon at approximately 3.5\%. The high proportion of neutral sentiment is notable given that the data comes from politically aligned channels. However, many posts in these channels function as news-style updates rather than direct expressions of opinion. A large portion of the collected messages consisted of event reporting, casualty updates, official statements, and descriptions of military developments. Such posts may contain strong political context without using language that a sentiment classifier identifies as positive or negative. The low positive sentiment rate follows a similar pattern. Explicitly positive language is uncommon in conflict-related reporting, even within politically aligned channels. Messages describing developments favourable to one side are often framed through reporting, justification, or criticism of the opposing side rather than through clearly positive emotional language. Negative sentiment was mainly associated with messages discussing casualties, humanitarian conditions, and conflict escalation. The results suggest that negative sentiment appears most consistently when posts describe losses or immediate consequences of violence, while developments favourable to one side are more often expressed through stance and framing rather than through explicitly positive sentiment (see Section~\ref{sec:framing-results}).

\begin{figure}[t]
\centering
\includegraphics[width=0.85\linewidth]{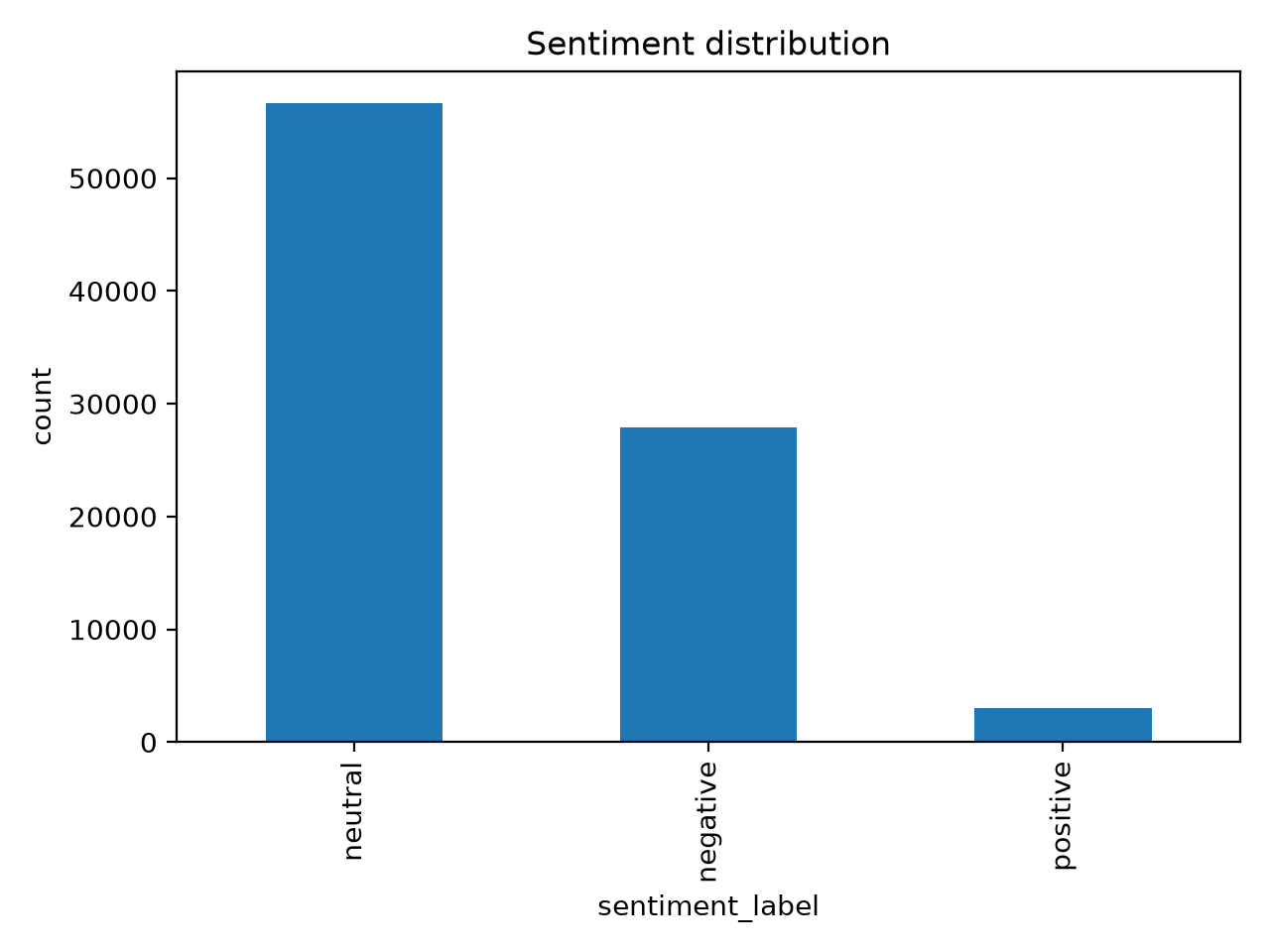}
\caption{Sentiment distribution across the full dataset.}
\label{fig:6_1}
\end{figure}

\subsection{Evaluation of Stance Detection Methods}
\label{sec:evaluation}
\begin{figure}[t]
\centering
\includegraphics[width=0.95\linewidth]{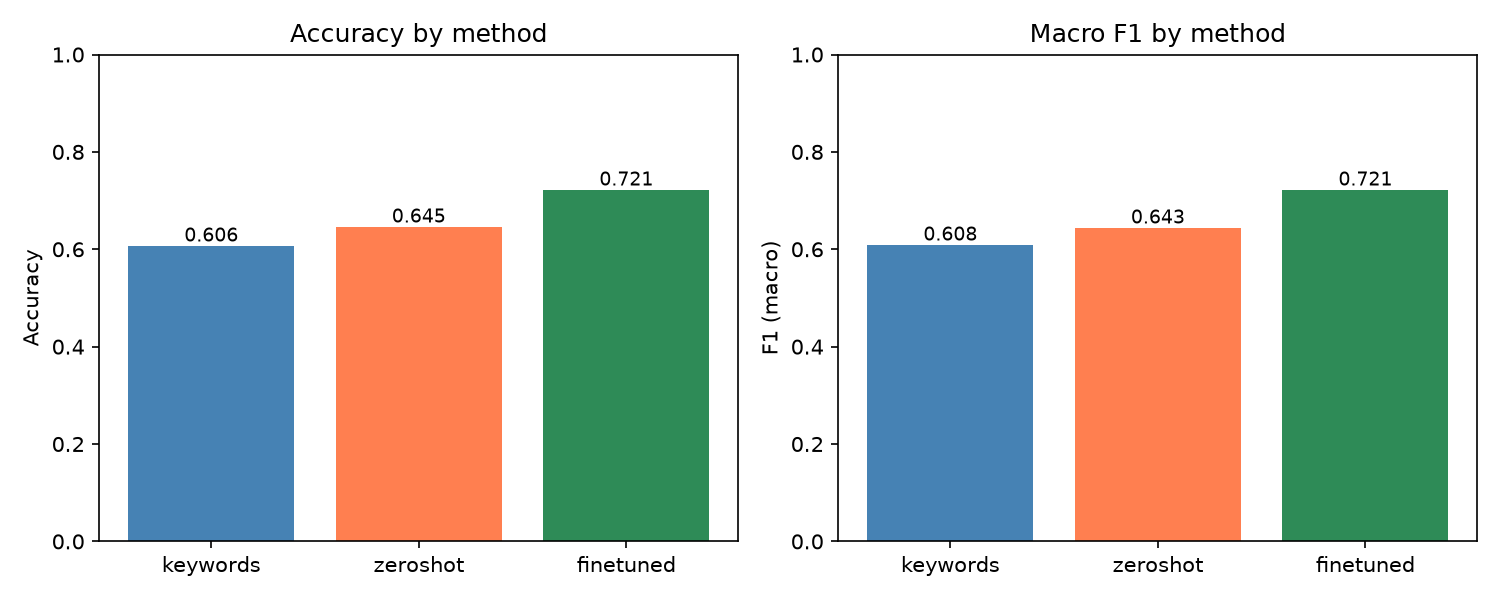}
\caption{Accuracy and macro F1 by stance-detection method.}
\label{fig:6_2}
\end{figure}
Figure~\ref{fig:6_2} shows accuracy and macro F1 scores for all three stance detection methods, each evaluated against the same 736 manually annotated messages. The fine-tuned BERTweet model achieved the highest performance, with a macro F1 of 0.721 and accuracy of 72.1\%, evaluated using 5-fold stratified cross-validation (F1 macro: 0.721 ± 0.027, accuracy: 0.721 ± 0.029 across folds). The keyword-based method achieved a macro F1 of 0.608 and accuracy of 60.6\%, while the zero-shot DeBERTa model achieved a macro F1 of 0.643 and accuracy of 64.5\%. All three methods were evaluated on the same 736 labeled rows, making the comparison directly fair. That keyword and zero-shot land within a few points of each other is notable given that the two approaches rely on fundamentally different mechanisms. The keyword approach uses manually constructed conflict-specific lexicons, while zero-shot DeBERTa uses natural language inference over a pre-trained transformer. Neither method was exposed to the Telegram-specific framing patterns present in this dataset, which likely explains why both stalled well below the fine-tuned model. The zero-shot model shows an interesting asymmetry: high precision for pro-Palestine (0.90) but low recall (0.44), indicating it identified pro-Palestine stance conservatively, only when the signal was strong, while missing a large portion of pro-Palestine messages.

\begin{figure}[t]
\centering
\includegraphics[width=0.85\linewidth]{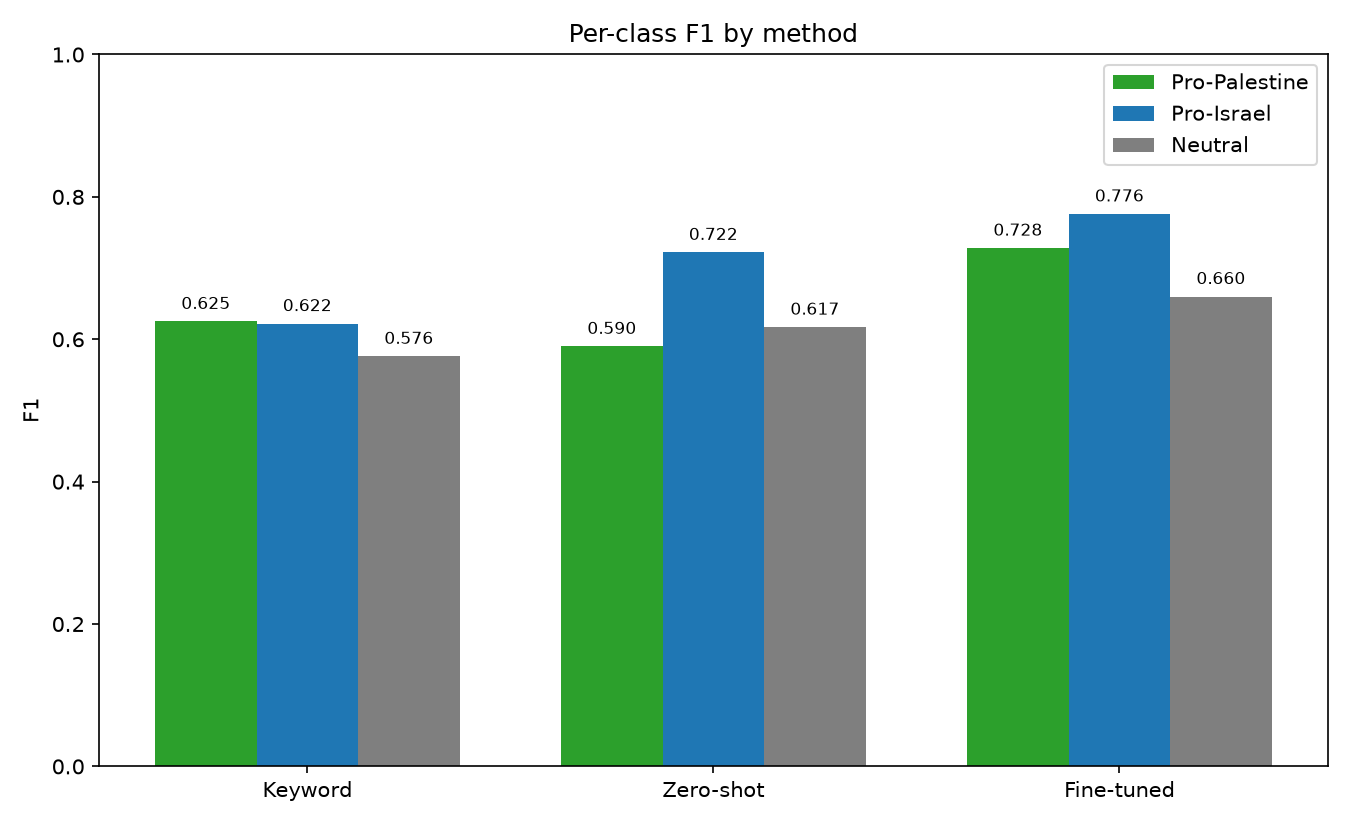}
\caption{Per-class F1 by stance-detection method.}
\label{fig:6_3}
\end{figure}
The fine-tuned model outperformed both baselines by roughly 8 to 11 percentage points on macro F1 (7.8 points over zero-shot, 11.3 over the keyword method). Per-class, the fine-tuned model performed strongest on pro-Israel (F1 0.78), followed by pro-Palestine (F1 0.73) and neutral (F1 0.66). As Figure~\ref{fig:6_3} shows, neutral is the hardest class for the keyword and fine-tuned methods, but for zero-shot the hardest class is instead pro-Palestine (F1 0.59, below its own neutral F1 of 0.62), a direct consequence of the very low pro-Palestine recall (0.44) noted above. BERTweet's pre-training on 850 million tweets provided a strong prior for informal social-media text, and fine-tuning on the 736 in-domain examples adapted it to the political vocabulary and stance distinctions specific to this dataset. Performance would likely improve further with a larger annotated sample; 736 labeled messages is a practical constraint, not a ceiling, consistent with the general finding in stance detection literature that fine-tuned in-domain models outperform zero-shot and lexicon-based approaches \cite{kucuk2020}.

\subsubsection{Keyword-Based Method}
\label{sec:eval-keyword}
The keyword method correctly classified 156 of 253 pro-Palestine messages, 134 of 253 pro-Israel messages, and 156 of 230 neutral messages. The dominant error pattern across all three classes was classification as neutral: 84 pro-Palestine and 72 pro-Israel messages received neutral labels, indicating that the lexicons failed to detect a usable stance signal in these cases rather than assigning the wrong one.

\begin{figure}[t]
\centering
\includegraphics[width=0.95\linewidth]{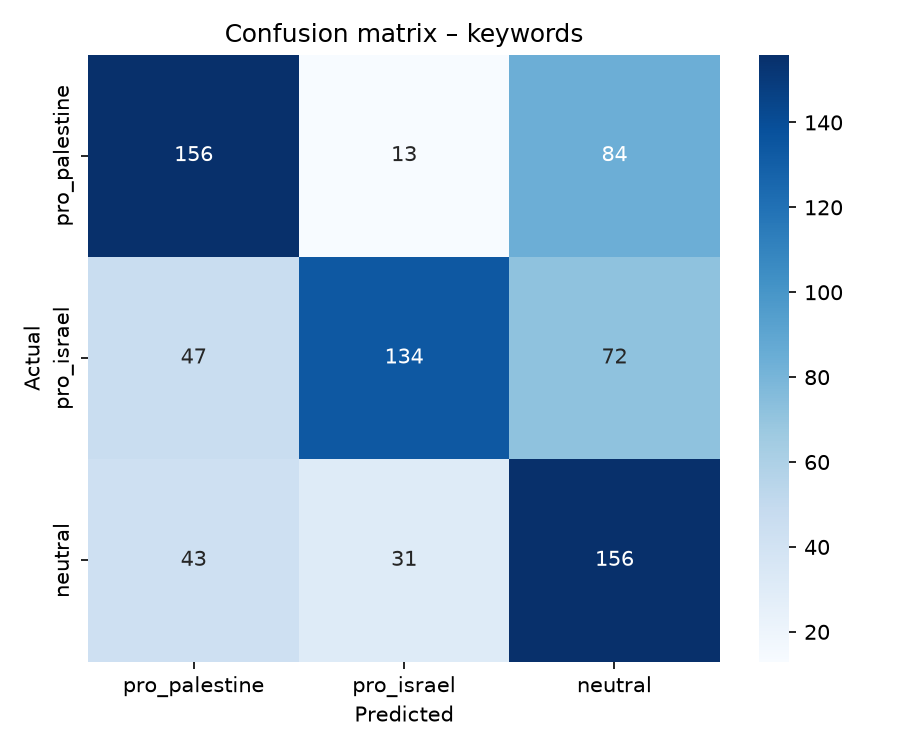}
\caption{Confusion matrix for the keyword-based stance method.}
\label{fig:6_4}
\end{figure}
Direct confusion between the two stance categories was, by contrast, uncommon. Only 13 pro-Palestine messages were classified as pro-Israel, and 47 pro-Israel messages as pro-Palestine. This asymmetry between frequent abstention (156 messages sent to neutral) and rare cross-class confusion (60 messages) reveals the method's fundamental operating characteristic: it is high-precision but low-recall. When the lexicons detected a clear stance signal, they usually assigned it to the correct side; their failure mode was an inability to detect a signal at all. This behaviour is structurally expected. A lexicon-based approach classifies as neutral any message where no listed term appears, or where matched terms from opposing lexicons cancel out. It cannot infer stance from context, syntactic structure, or argument. Two distinct mechanisms drive the errors observed here. The first is term weighting. Several pro-Palestine geographic markers, "Gaza", "Rafah", "West Bank", are present only in the pro-Palestine lexicon, but at the lowest weight. When such a low-weight term co-occurs with a higher-weight pro-Israel term, the pro-Israel signal dominates or the two offset into a neutral result. Pro-Israel keywords, by comparison, are more often tied to specific events or narratives, "October 7 massacre", "Nova festival", "bring them home", which carry high weight and appear rarely in opposing-side content, producing cleaner matches when present. This contributes to pro-Palestine messages being classified as neutral more often than pro-Israel messages. The second mechanism is a token-stance mismatch that the lexicon cannot resolve, and it produces the method's most revealing errors. Consider a message reporting that large quantities of aid sit undelivered at the Kerem Shalom crossing into Gaza, attributing the failure to the United Nations and to Hamas, and concluding that the humanitarian crisis is manufactured. A human annotator labelled this pro-Israel, since its argument blames Hamas and disputes the severity of the crisis. The lexicon, however, matched the pro-Palestine geographic and humanitarian terms ("Gaza", crisis-related vocabulary) while finding no term that captures the message's actual pro-Israel argument, which is carried entirely by structure and assertion rather than by any listed keyword. The message was therefore assigned to the opposing side. Such cases illustrate the ceiling on a purely lexical approach: accuracy on messages with implicit or argumentative stance is limited not by classification error but by coverage of the surface vocabulary alone.

\subsubsection{Zero-Shot (DeBERTa)}
\label{sec:eval-zeroshot}
The zero-shot method correctly classified 111 of 253 pro-Palestine messages, 178 of 253 pro-Israel messages, and 186 of 230 neutral messages. As with the keyword method, the dominant error was abstention into the neutral class rather than confusion between the two stance categories: 118 of 253 pro-Palestine messages received a neutral label, producing a recall of just 43.9\% for this class, while pro-Israel and neutral recall were considerably higher at 70.4\% and 80.9\% respectively.

\begin{figure}[t]
\centering
\includegraphics[width=0.95\linewidth]{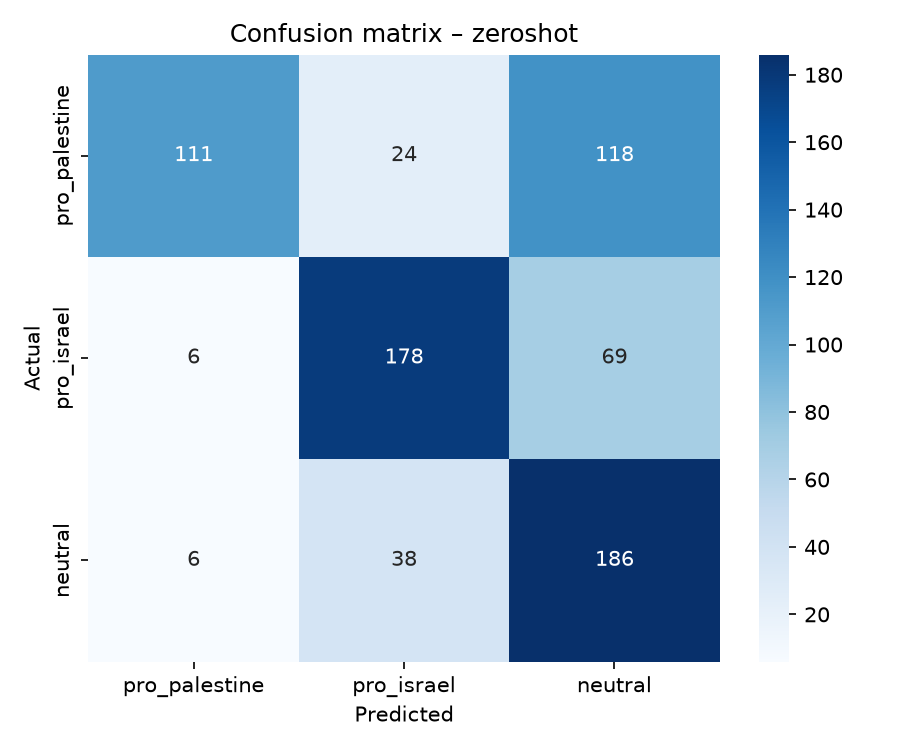}
\caption{Confusion matrix for the zero-shot DeBERTa method.}
\label{fig:6_5}
\end{figure}
The structural similarity to the keyword method is worth noting. Direct confusion between the two explicit stance classes was minimal, only 24 pro-Palestine messages were classified as pro-Israel and 6 pro-Israel as pro-Palestine. Like the lexicon approach, the zero-shot model rarely takes the wrong side; its errors consist almost entirely of falling back on the neutral label when no strong signal is present. Both non-fine-tuned methods therefore share the same failure mode, differing only in degree, and the pro-Palestine class is where that failure is most pronounced. The model was not trained on any data from this dataset and had no exposure to the specific vocabulary, framing conventions, or political terminology of the collected channels. Zero-shot classification with DeBERTa operates through natural language inference, where each message is evaluated against candidate labels and assigned the one with the highest entailment score \cite{laurer2024}. Performance therefore depends on how closely the input language aligns with patterns learned during pretraining. This dependence explains the pro-Palestine recall gap directly. Pro-Israel messages more frequently contain language that overlaps with widely represented conflict discourse in general English corpora, references to terrorism, hostages, and military operations, which entail the pro-Israel label readily. A large portion of the pro-Palestine class, by contrast, consists of regional-actor framing carried by Iran, Hezbollah, and Houthi sources rather than by Palestine-specific vocabulary. A representative misclassified example is a message reporting that Iranian air defences downed a hostile reconnaissance drone: an annotator reads it as pro-Palestine within the wider axis-of-resistance framing of the channel, but the message contains no term and no proposition that entails a "pro-Palestine" stance in general English, and the model assigns it to neutral. The same fate befalls formal military and political statements from these actors, which read as factual reportage to a model without the regional context. These forms of expression produce weak entailment scores, which is the proximate cause of the elevated neutral misclassification rate for pro-Palestine content relative to pro-Israel content.

\subsubsection{Fine-Tuned (BERTweet)}
\label{sec:eval-finetuned}
The fine-tuned model achieved the strongest results across all three classes. Pro-Palestine recall reached 75.1\% (190 of 253 correct), and pro-Israel recall was 73.1\% (185 of 253 correct). The neutral class, while the weakest, still reached 67.8\% recall (156 of 230 correct), an improvement over both baseline methods.

\begin{figure}[t]
\centering
\includegraphics[width=0.95\linewidth]{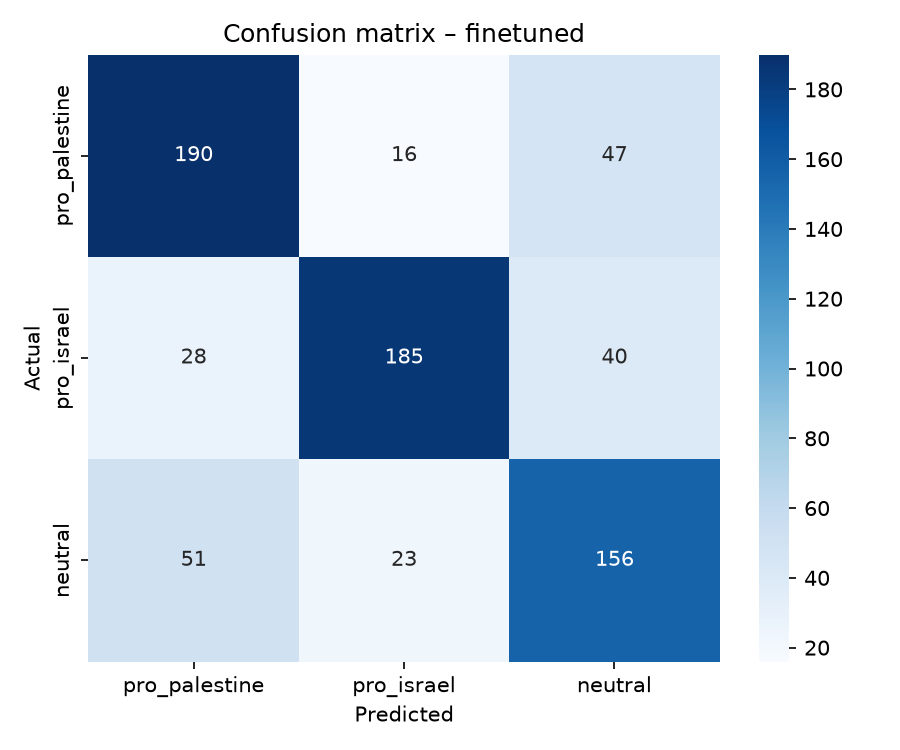}
\caption{Confusion matrix for the fine-tuned BERTweet method.}
\label{fig:6_6}
\end{figure}
The performance on the explicit stance classes reflects the benefit of fine-tuning on domain-specific labeled data. The model was trained on examples drawn directly from the same Telegram channels and conflict context as the evaluation set, allowing it to learn the vocabulary, framing patterns, and contextual signals that distinguish pro-Palestine from pro-Israel content in this dataset. Crucially, this is the same regional-actor and argumentative content that defeated the keyword and zero-shot methods: by learning from in-domain examples, the fine-tuned model recovers much of the stance signal that the other two approaches abstained on, which is the principal source of its recall gain on the pro-Palestine class. This reduces reliance on general-purpose semantic similarity and improves sensitivity to domain-specific expressions of stance. This gain, however, comes with a mild directional bias that the confusion matrix makes visible. The model's errors are not symmetric: it assigns 269 messages to the pro-Palestine class against a true support of 253, while assigning only 224 to pro-Israel against a true support of 253. Both dominant off-diagonal errors feed the pro-Palestine class, 51 of the 74 misclassified neutral messages and 28 of the 68 misclassified pro-Israel messages were labelled pro-Palestine. The model therefore over-predicts pro-Palestine and under-predicts pro-Israel, a tendency likely inherited from the framing patterns most strongly represented in the training data. The effect is modest and does not undermine the per-class results, but it should be considered when interpreting the full-dataset distribution in Section~\ref{sec:distribution}, where any systematic tilt is applied at scale. The neutral class is difficult for every method in this study, the hardest class for the keyword and fine-tuned methods, and second-hardest behind pro-Palestine for zero-shot (Section~\ref{sec:eval-zeroshot}), consistent with the annotation-boundary difficulty documented in the stance literature \cite{kucuk2020}. Messages labelled as neutral often still carry weak or indirect stance signals, blurring the boundary between neutral and non-neutral, a difficulty compounded by the known unreliability of annotation at this boundary \cite{kucuk2020}. The dataset was annotated by a single annotator over 736 messages, so inter-annotator agreement could not be computed; annotation reliability is known to degrade for ambiguous categories, where the ambiguity of the data and the labelling task directly lowers the agreement coders reach \cite{nowak2010}. This affects the neutral class more strongly than the two explicit stance classes, where the linguistic signals are clearer and more consistent. Taken together, the three methods form a consistent progression. The keyword and zero-shot approaches share a single failure mode, abstention into neutral when no surface signal is present, and differ only in how often they fall into it. Fine-tuning addresses that failure directly, converting previously undetected stance into correct classification, at the cost of a mild pro-Palestine lean in its residual errors. All performance figures are based on leakage-free out-of-fold predictions across the 5-fold cross-validation, ensuring that no labelled message was evaluated by a model that had seen it during training.

\subsection{Inter-Method Agreement}
\label{sec:agreement}
Inter-method agreement was calculated across the full 87,617-message dataset by comparing the stance labels assigned by each pair of methods. Keywords and zero-shot agreed on 61.3\% of messages, keywords and fine-tuned on 60.2\%, and zero-shot and fine-tuned on 62.7\%. In all three comparisons, roughly 40\% of messages received a different label depending on the method used.

\begin{figure}[t]
\centering
\includegraphics[width=\linewidth]{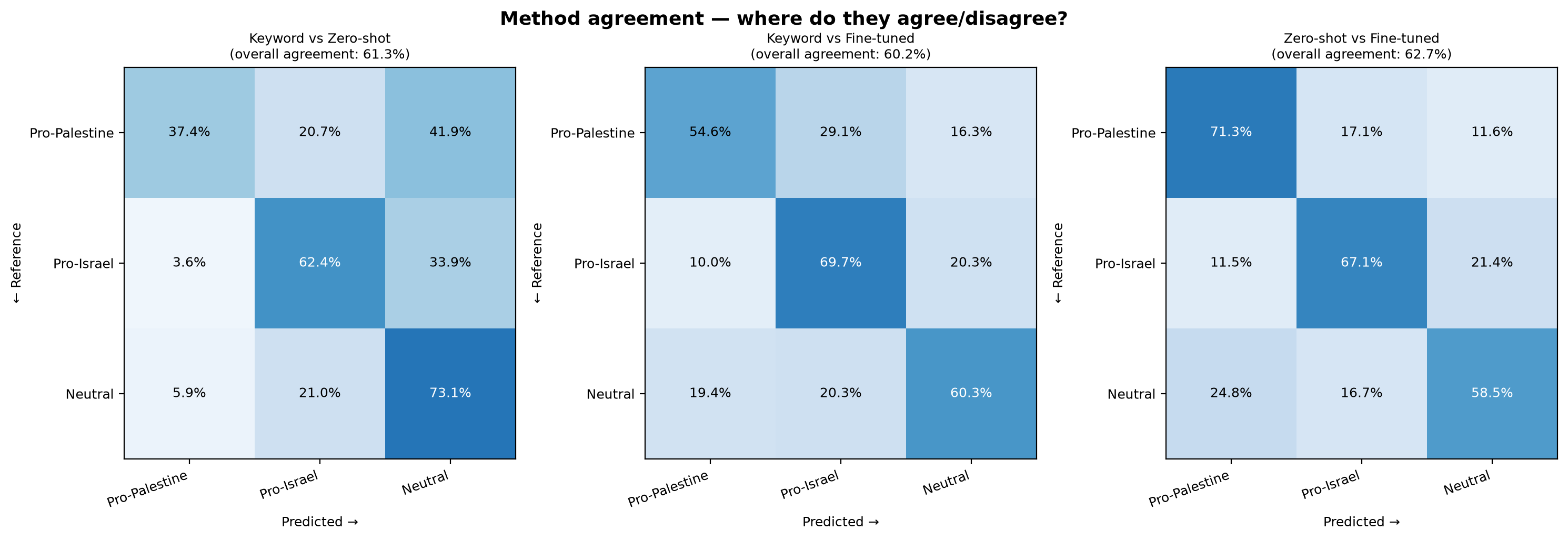}
\caption{Pairwise inter-method agreement (row-normalised).}
\label{fig:6_7}
\end{figure}
The per-class panels in Figure~\ref{fig:6_7} are row-normalised: each row shows how one method's assignments of a given stance were distributed by the other method. Read this way, the disagreements are concentrated on the pro-Palestine class for the two pairs involving the keyword method. When keywords assigned a pro-Palestine label, the zero-shot model agreed only 37.4\% of the time and reassigned 41.9\% of those messages to neutral; the fine-tuned model agreed with keywords on pro-Palestine content in 54.6\% of cases. This matches the evaluation results: the keyword lexicons fire on surface pro-Palestine vocabulary that the other two methods, which weigh context, frequently read as neutral. The pattern reverses for the zero-shot and fine-tuned pair, which agreed on 71.3\% of the messages zero-shot labelled pro-Palestine, the highest pro-Palestine concordance of any pair. This should be read in light of zero-shot's conservatism established in Section~\ref{sec:eval-zeroshot}: because the model assigns a pro-Palestine label only when the signal is strong, its pro-Palestine set is a small, high-confidence subset that the fine-tuned model agrees with readily. The two methods therefore concur most on the clearest pro-Palestine content, rather than across the class as a whole. Across all three method pairs, disagreement was far more likely to take the form of one method assigning a neutral label than of a direct conflict between the two opposing stance categories; pro-Palestine versus pro-Israel disagreements were rare. This mirrors the per-method error analysis in Section~\ref{sec:evaluation}, where the default to neutral, not cross-stance confusion, was the dominant failure mode of every method.

\subsection{Stance Distribution Across the Full Dataset}
\label{sec:distribution}
The three methods produce noticeably different distributions. The keyword and zero-shot methods classify the majority of the dataset as neutral, 55.9\% and 58.1\% respectively, while the fine-tuned model assigns a smaller neutral share at 41.5\%. The fine-tuned model assigns 27.9\% of messages as pro-Palestine and 30.6\% as pro-Israel, producing the most balanced distribution across the two explicit stance classes. The keyword method assigns 28.3\% pro-Palestine but only 15.8\% pro-Israel, while zero-shot assigns 14.5\% pro-Palestine and 27.4\% pro-Israel.

\begin{figure}[t]
\centering
\includegraphics[width=\linewidth]{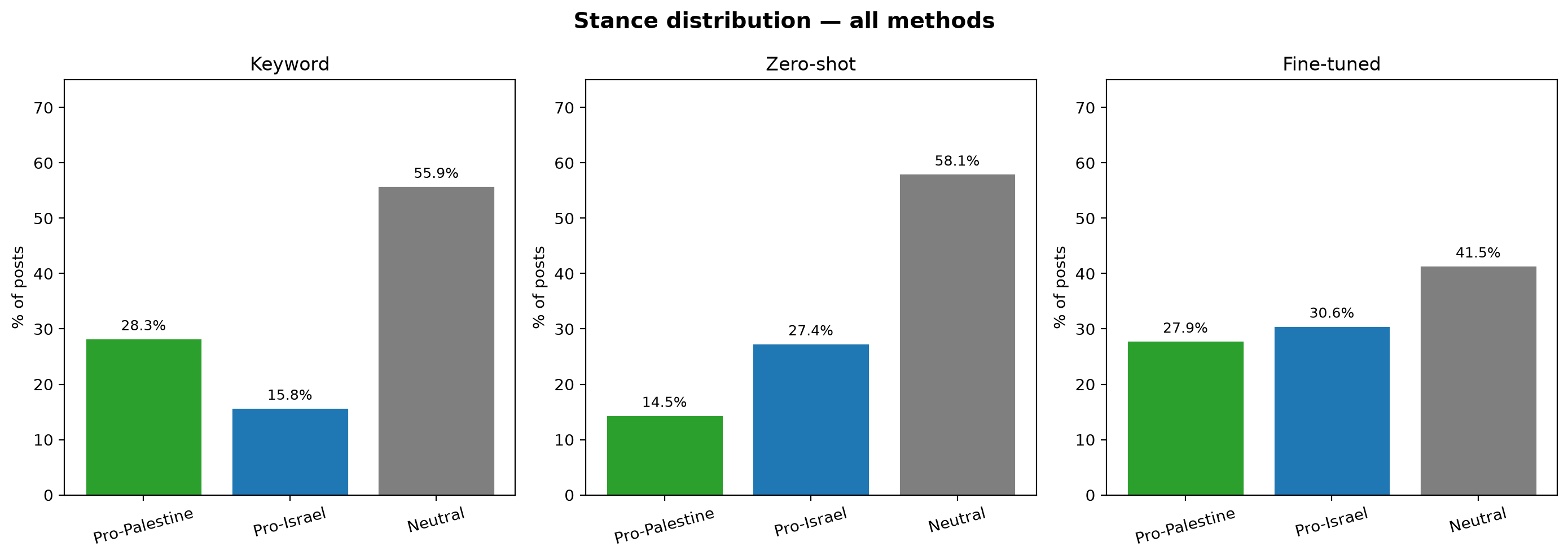}
\caption{Stance label distribution across the full corpus by method.}
\label{fig:6_8}
\end{figure}
This difference reflects how each method handles implicit stance signals. Keywords require explicit lexicon matches and default to neutral when none are found. Zero-shot relies on general semantic alignment with label descriptions and similarly produces low confidence on messages without clear political markers. The fine-tuned model, trained on manually labelled examples from the same dataset, learned that stance is often expressed through framing, context, and indirect language rather than explicit vocabulary alone. As a result, it assigns politically aligned labels more frequently and reduces the neutral share considerably, which is the full-dataset manifestation of the recall gain documented in Section~\ref{sec:evaluation}. The direction of the differences tracks the evaluation results. Zero-shot achieved the lowest recall for pro-Palestine messages and correspondingly produced the smallest pro-Palestine share in the full dataset. The strongest evidence that this reflects a measurement artifact rather than a real scarcity of pro-Palestine content is internal to zero-shot's own output: it assigns nearly twice as many messages to pro-Israel (27.4\%) as to pro-Palestine (14.5\%), despite the corpus being collected in approximately equal volume from the two stance groups (roughly 47,000 messages each before cleaning, Section~\ref{sec:data-collection}). This asymmetry aligns directly with the model's low pro-Palestine recall established in Section~\ref{sec:eval-zeroshot}, where a large share of genuinely pro-Palestine messages, particularly regional-actor framing, was misclassified as neutral. Notably, the two label-free methods skew in opposite directions: zero-shot under-assigns pro-Palestine (14.5\% against 27.4\% pro-Israel), while the keyword method under-assigns pro-Israel (15.8\% against 28.3\% pro-Palestine). Because the baselines err toward opposite classes, their full-dataset distributions cannot be reconciled with each other, and only the fine-tuned model yields a balanced split (27.9\% versus 30.6\%) consistent with the approximately equal collection volumes. This makes the fine-tuned distribution the only one of the three that can be read as an estimate of the corpus composition rather than an artifact of method bias. Finally, the fine-tuned model's distribution warrants a note in light of the directional error tendency identified in Section~\ref{sec:eval-finetuned}. Although the model slightly over-predicted pro-Palestine on the labelled evaluation set, this tendency is mild and does not produce a pro-Palestine-skewed output at scale: across the full dataset it assigns marginally more messages to pro-Israel (30.6\%) than to pro-Palestine (27.9\%). The two explicit stance classes are therefore close to balanced in the model's output, with the residual evaluation-set lean too small to materially distort the full-dataset proportions.

\subsection{Sentiment by Stance-All Methods}
\label{sec:sentiment-by-stance}
\begin{figure}[t]
\centering
\includegraphics[width=\linewidth]{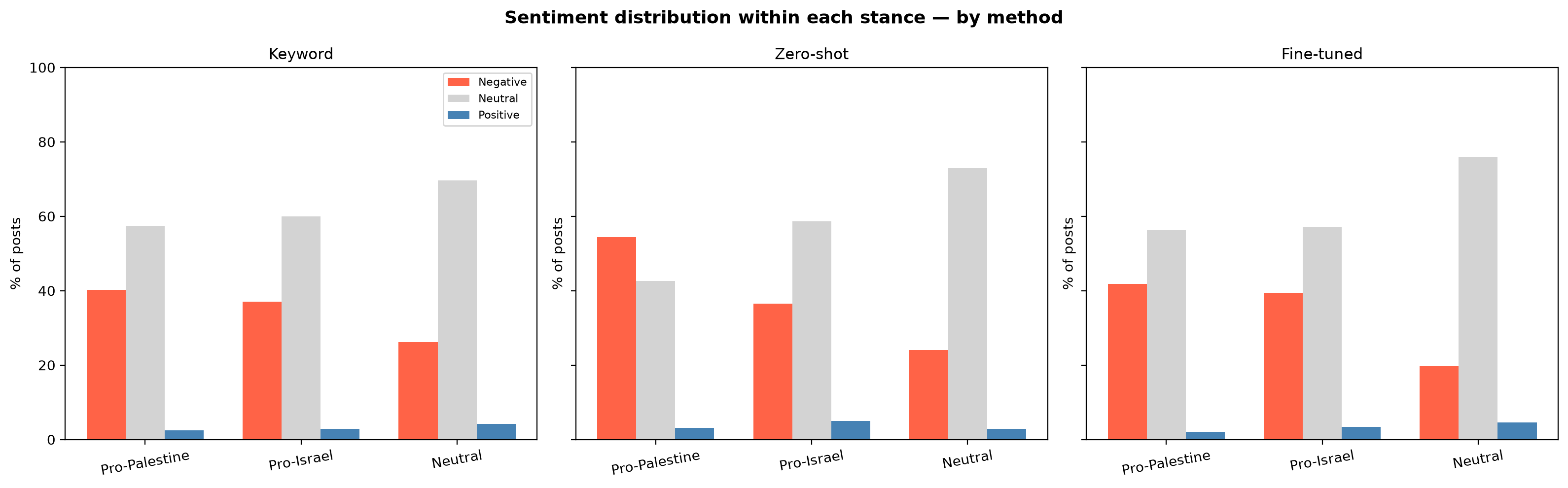}
\caption{Stance--sentiment relationship across methods.}
\label{fig:6_9}
\end{figure}
Figure~\ref{fig:6_9} shows that the broad relationship between stance and sentiment is consistent across all three stance detection methods. In every panel, neutral-stance messages contain the highest proportion of neutral sentiment and the lowest proportion of negative sentiment, while messages expressing an explicit political stance, whether pro-Palestine or pro-Israel, carry more negative sentiment. Positive sentiment remains rare across all categories, never exceeding roughly 5\%. The stability of this pattern across three methodologically different classifiers suggests the stance--sentiment relationship is a property of the data rather than an artifact of any single classification approach. Within this shared pattern, the two explicit stance classes are not symmetric. Pro-Palestine messages carry a higher share of negative sentiment than pro-Israel messages under all three methods. Aggregated across every message in a stance group the gap is contained, most visibly under zero-shot, where pro-Palestine is the only stance for which negative sentiment exceeds neutral sentiment (approximately 54\% versus 42\%). It is substantially larger, however, once messages are restricted to the death- and victim-framing subsets analysed in Section~\ref{sec:framing-results} (pro-Palestine 62\% versus pro-Israel 45\% for death context, and 58\% versus 35\% for victim framing), where the difference in emotional register between the two groups is concentrated. This asymmetry, pro-Palestine content skewing more negative than pro-Israel content of the same stance strength, anticipates the framing analysis in Section~\ref{sec:framing-results}, where the two groups are shown to discuss comparable events in markedly different emotional registers.

\subsection{Framing Analysis}
\label{sec:framing-results}
Framing rates in this section are computed within each stance group as assigned by the fine-tuned model, that is, the percentage of a group's messages containing each framing category, and exclude messages the model classified as stance-neutral (the sentiment of the remaining messages is then reported separately within each framing category). Across all four framing categories, military framing is the most common, appearing in 51.8\% of pro-Israel messages and 41.1\% of pro-Palestine messages. This is expected: the conflict is at its core a military one, and both sides discuss operations, weapons, and armed actors. Legal and international-law framing is the rarest category in both groups, though pro-Israel messages use it slightly more (approximately 6.3\% versus 3.2\%). All four framing-usage differences between the two stance groups are statistically significant (chi-square tests of independence, p < 0.001), as are the sentiment differences within each framing category reported below (p < 0.001). These framing categories are constructed through keyword matching and therefore carry the same context-insensitivity established for the keyword stance method in Section~\ref{sec:eval-keyword}: they measure the rate at which each group uses a given vocabulary, not a validated semantic judgment of how that vocabulary is deployed. Because these groups are defined by the model's predicted stance rather than the channel's source label, the framing comparison inherits the mild directional bias noted in Section~\ref{sec:eval-finetuned}; the effect is small relative to the framing differences observed here.

\begin{figure}[t]
\centering
\includegraphics[width=\linewidth]{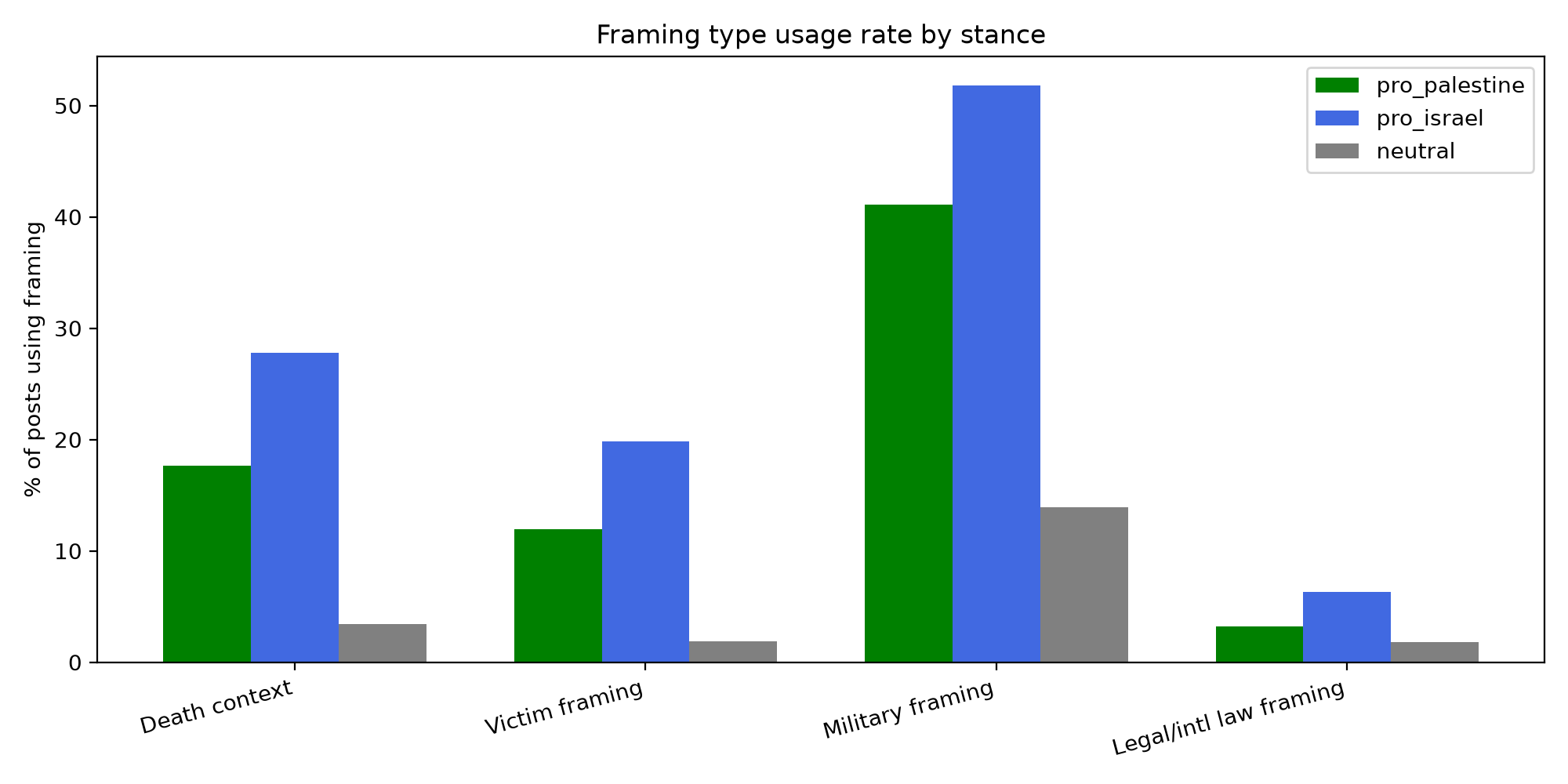}
\caption{Framing-category rates by fine-tuned stance group.}
\label{fig:6_10}
\end{figure}
\begin{figure}[t]
\centering
\includegraphics[width=\linewidth]{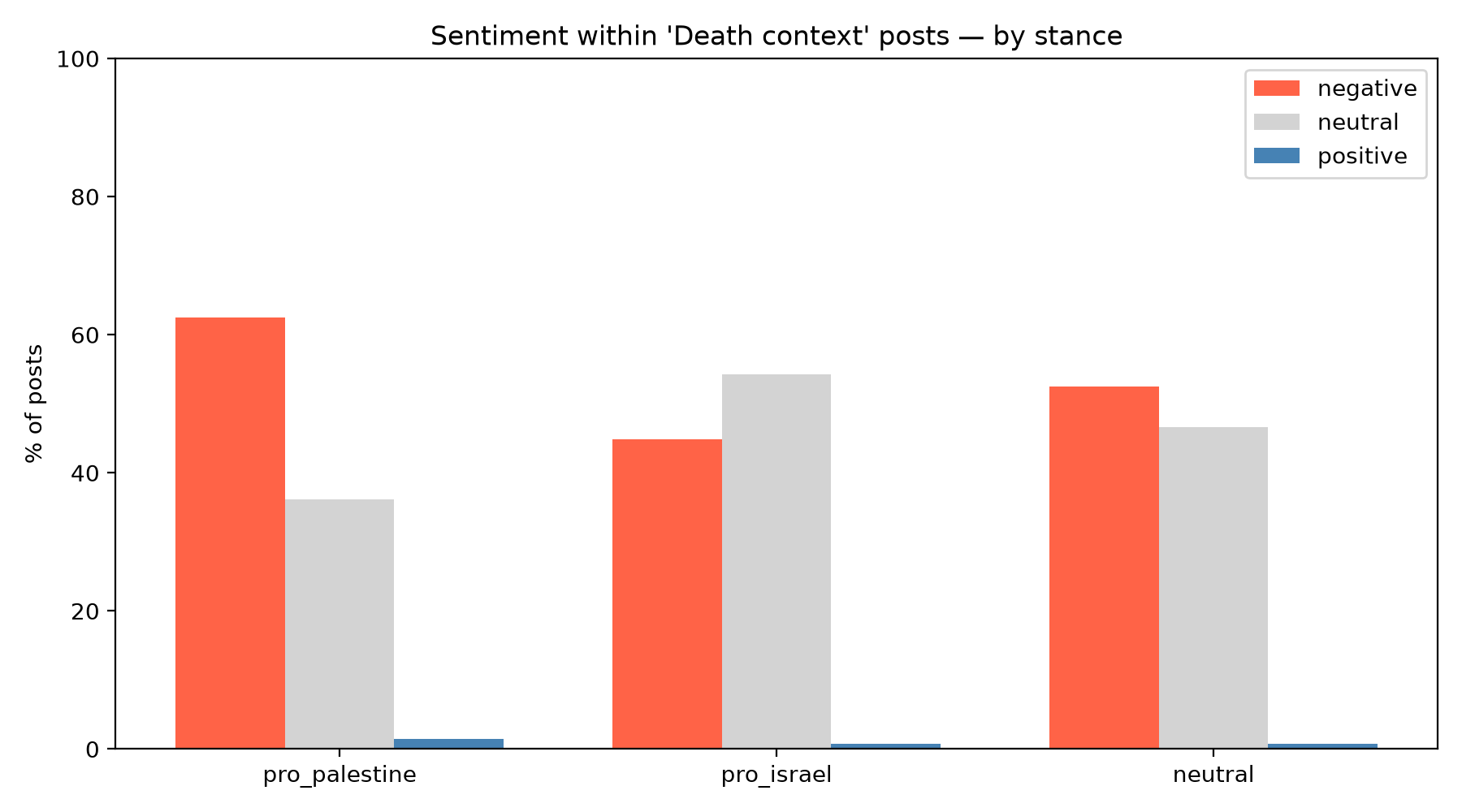}
\caption{Sentiment within death-context framing by stance group.}
\label{fig:6_11}
\end{figure}
\begin{figure}[t]
\centering
\includegraphics[width=\linewidth]{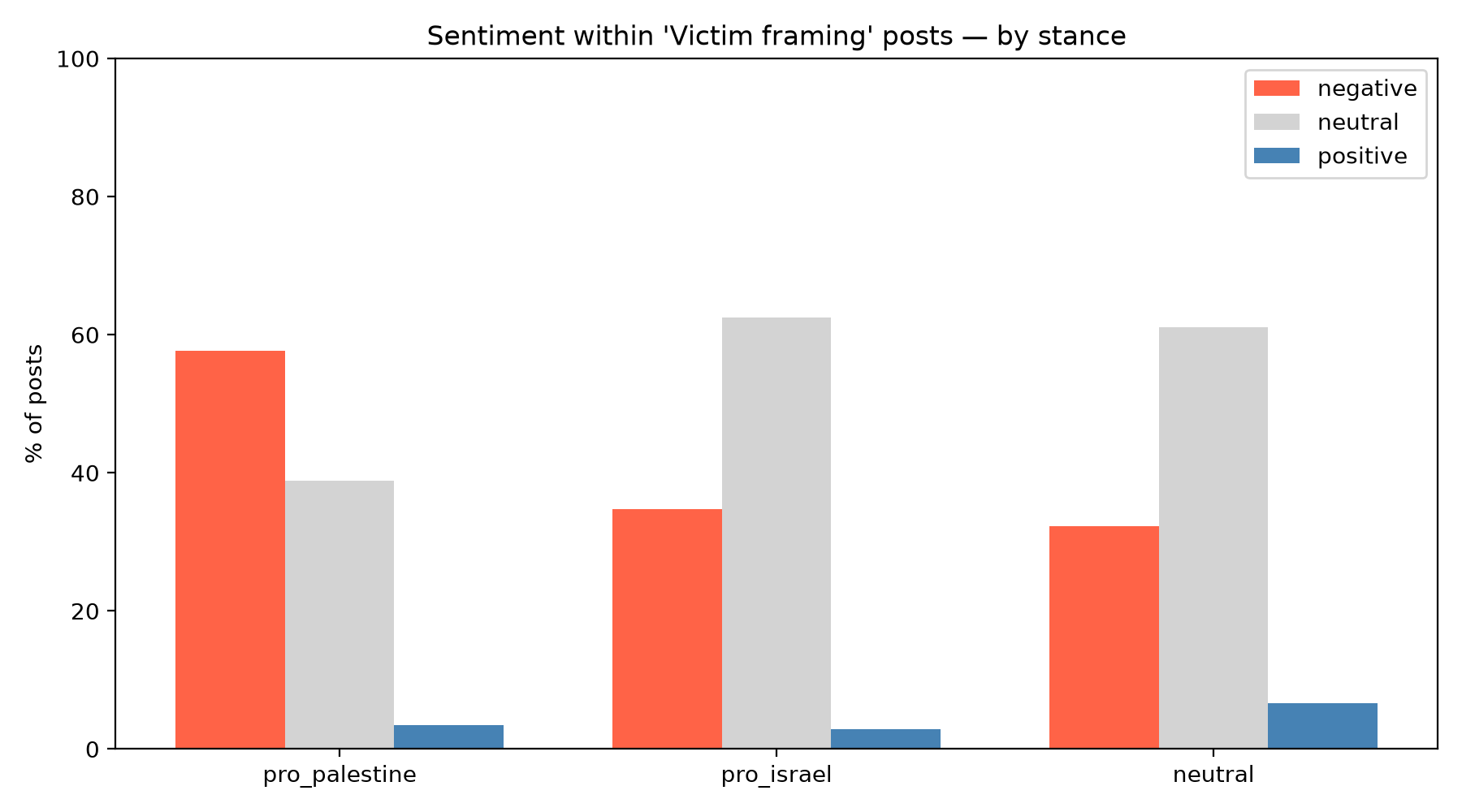}
\caption{Sentiment within victim framing by stance group.}
\label{fig:6_12}
\end{figure}

Less expected is that pro-Israel messages invoke death-context framing at a higher rate (27.8\%) than pro-Palestine messages (17.7\%), with the same direction holding for victim framing (19.8\% versus 11.9\%). The significance of this becomes clear only when the sentiment within these framings is considered (Figures~\ref{fig:6_11} and~\ref{fig:6_12}).

\begin{figure}[t]
\centering
\includegraphics[width=\linewidth]{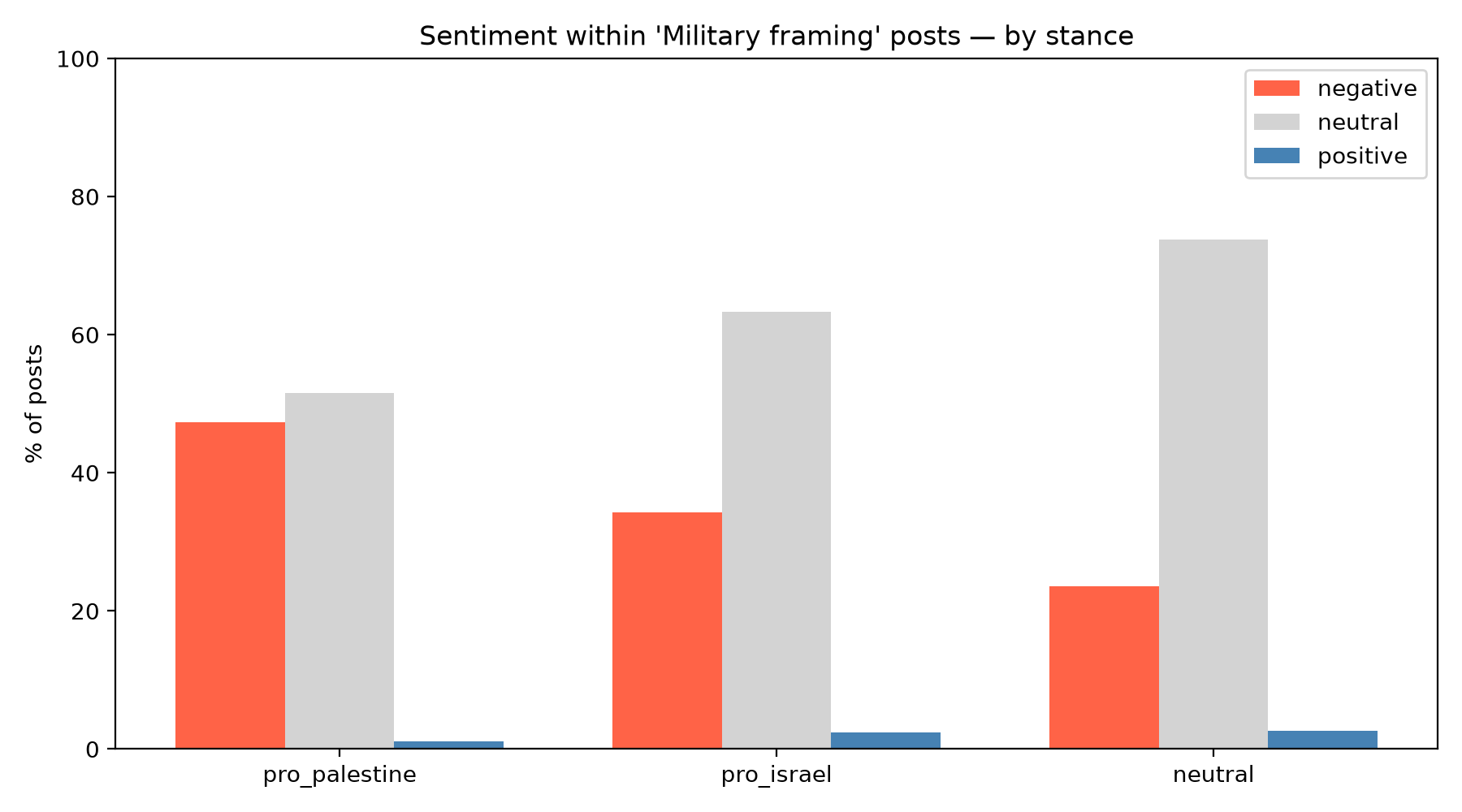}
\caption{Sentiment within military framing by stance group.}
\label{fig:6_13}
\end{figure}

\begin{figure}[t]
\centering
\includegraphics[width=\linewidth]{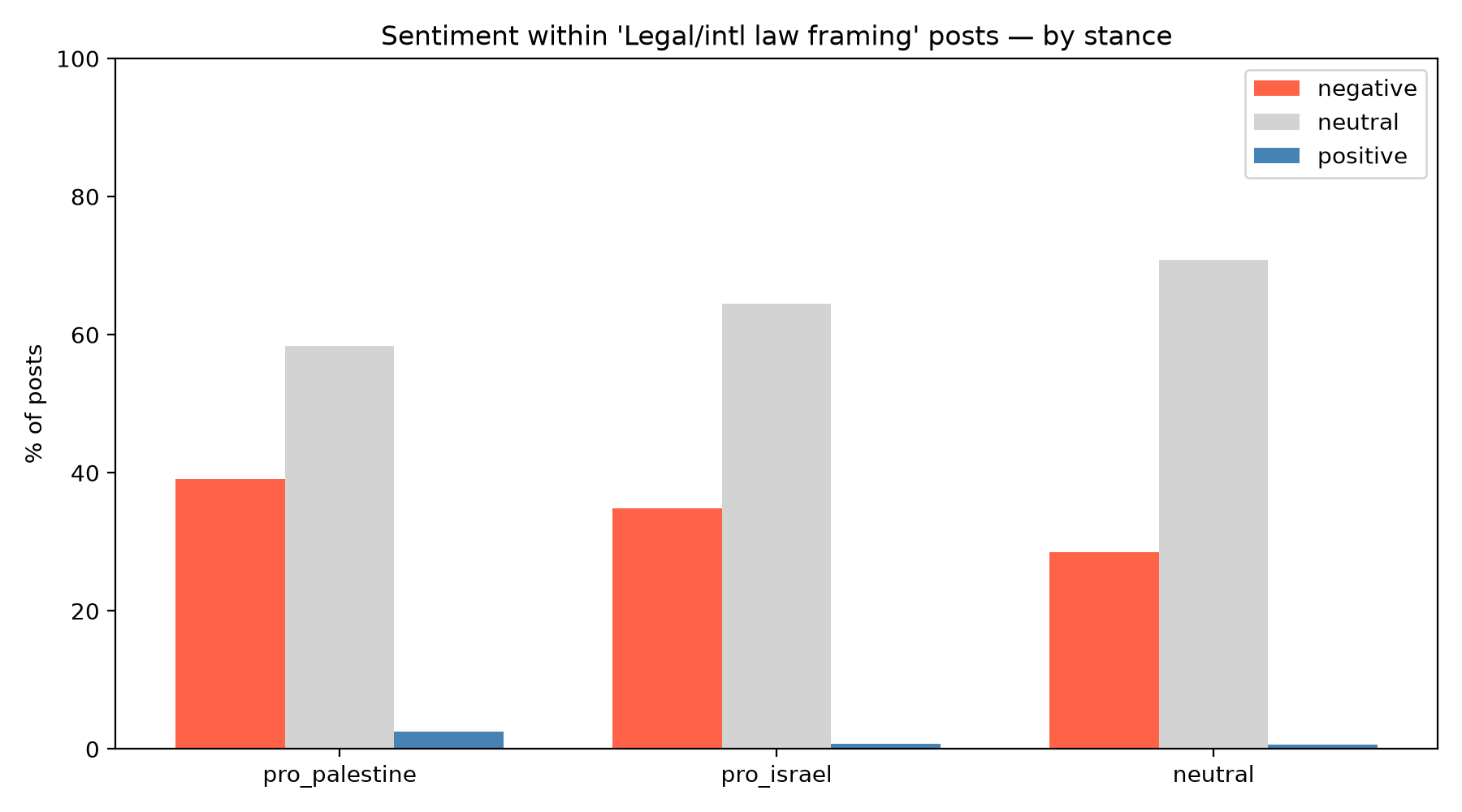}
\caption{Sentiment within legal framing by stance group.}
\label{fig:6_14}
\end{figure}

Here the relationship reverses. Although pro-Israel messages use death and victim framing more often, the emotional charge of those messages is consistently lower. In death-context posts, pro-Israel messages are predominantly neutral in sentiment (approximately 54\%, against 45\% negative), whereas pro-Palestine death-context posts are majority-negative (approximately 62\%). The gap is sharper for victim framing: pro-Palestine messages carry roughly 58\% negative sentiment compared to approximately 35\% for pro-Israel. The same vocabulary of death and victimhood is therefore deployed in two distinct registers, invoked more frequently by one side, but with far greater emotional weight by the other. This contrast reflects the position each group occupies relative to the events it describes. Pro-Israel channels write predominantly as the acting party: they report operations and their outcomes, describe military developments, and justify actions. When death appears in this content, it is typically the reported consequence of an operation rather than a loss suffered by the channel's own side, and a factual, report-style register is therefore the expected mode, which is exactly what the high usage combined with predominantly neutral sentiment shows. Pro-Palestine channels, by contrast, more often write as the affected party, invoking the same vocabulary of death and victimhood to describe losses to their own side, which is naturally expressed through negative sentiment. The asymmetry is thus not a contradiction but a direct signature of the two discourse positions: the actor describing outcomes, and the affected party describing harm. (Sentiment here is inferred from the classifier's labels and reflects measurable differences in sentiment-bearing language.) Military framing (Figure~\ref{fig:6_13}) shows a smaller but consistent version of the same pattern: pro-Palestine military messages carry higher negative sentiment (approximately 47\%) than pro-Israel (approximately 34\%), suggesting the difference in register extends beyond death and victim framing into operational discussion. Legal framing (Figure~\ref{fig:6_14}) shows similar negative-sentiment rates across both groups (approximately 39\% versus 35\%) and, given its rarity in the corpus, supports no firm conclusion.


\subsection{Event Analysis}
\label{sec:events}
Event analysis was conducted around five major conflict-related milestones using a ±14-day window. These five events were selected because they had sufficient message volume for meaningful day-level comparison; the remaining candidate events fell within sparsely covered periods of the corpus and are not examined here. Due to the reverse-chronological collection method described in Section~\ref{sec:data-collection}, earlier events contain fewer messages within the analysis window and are interpreted with more caution. Sentiment change around each event is described as each group's share of negative-sentiment messages in the fourteen days before the event, in days 0--7, and in days 8--14. The comparisons in this section are reported descriptively; given the number of events, groups, and windows involved, per-event significance tests are not reported, and formal testing is confined to the corpus-level framing comparisons of Section~\ref{sec:framing-results}. One limitation applies throughout. The dataset contains no neutral channels: every message originates from a pro-Palestine or pro-Israel source. Neutral labels refer only to individual messages classified by the fine-tuned model as lacking explicit stance markers; they do not represent a separate community. When neutral-labelled volume rises, this reflects messages from politically aligned channels that contain less explicit political framing, typically factual or report-style content. Across the military events analysed, the rise in neutral-sentiment, came predominantly from pro-Israel-source channels. Iran Operation True Promise 2 - October 1, 2024 On October 1, 2024, Iran launched approximately 180 ballistic missiles at Israel. Message volume rose sharply around the event date, but the immediate increase was led by neutral and pro-Israel content (85 and 61 messages on the event day, against 38 pro-Palestine). Pro-Palestine volume peaked several days later rather than on the date itself. The sentiment response was mirror-imaged. Pro-Israel negative sentiment rose from 30\% before the attack to 45\% in the first week and remained elevated at 43\% in the second. Pro-Palestine negativity moved the opposite way, falling from 31\% to 22\% in the first week before returning to 38\% in the second, close to its pre-event level. The two groups' first-week shares, 45\% against 22\%, moved twenty-three points apart: the clearest single instance in the dataset of the two communities responding to the same days of the same escalation in opposite emotional directions, consistent with a one-sided strike that harmed one community's side while marking a success for the other's.

\begin{figure}[t]
\centering
\includegraphics[width=\linewidth]{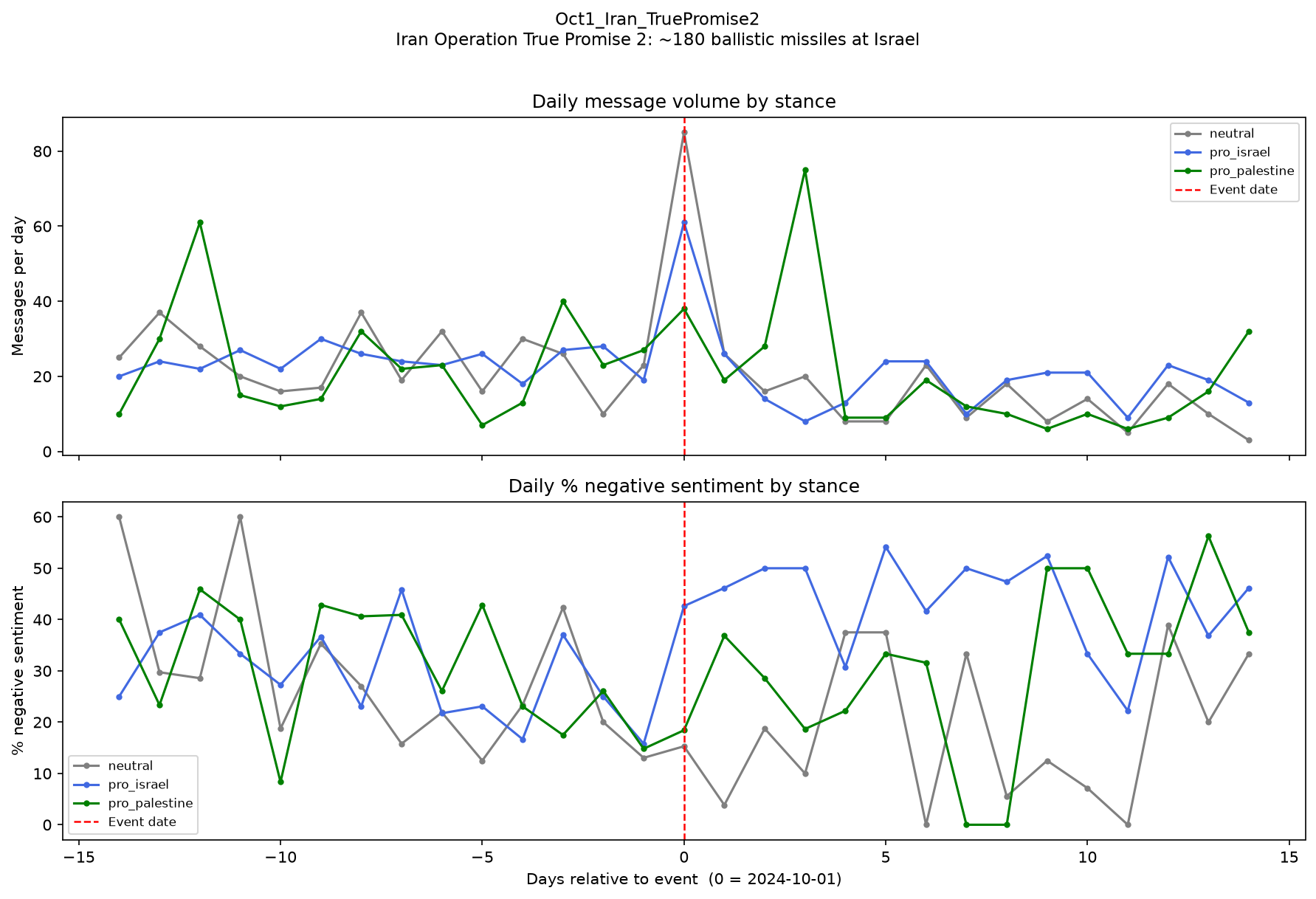}
\caption{Event window: Iran Operation True Promise 2 (1 Oct 2024).}
\label{fig:6_15}
\end{figure}
\begin{figure}[t]
\centering
\includegraphics[width=\linewidth]{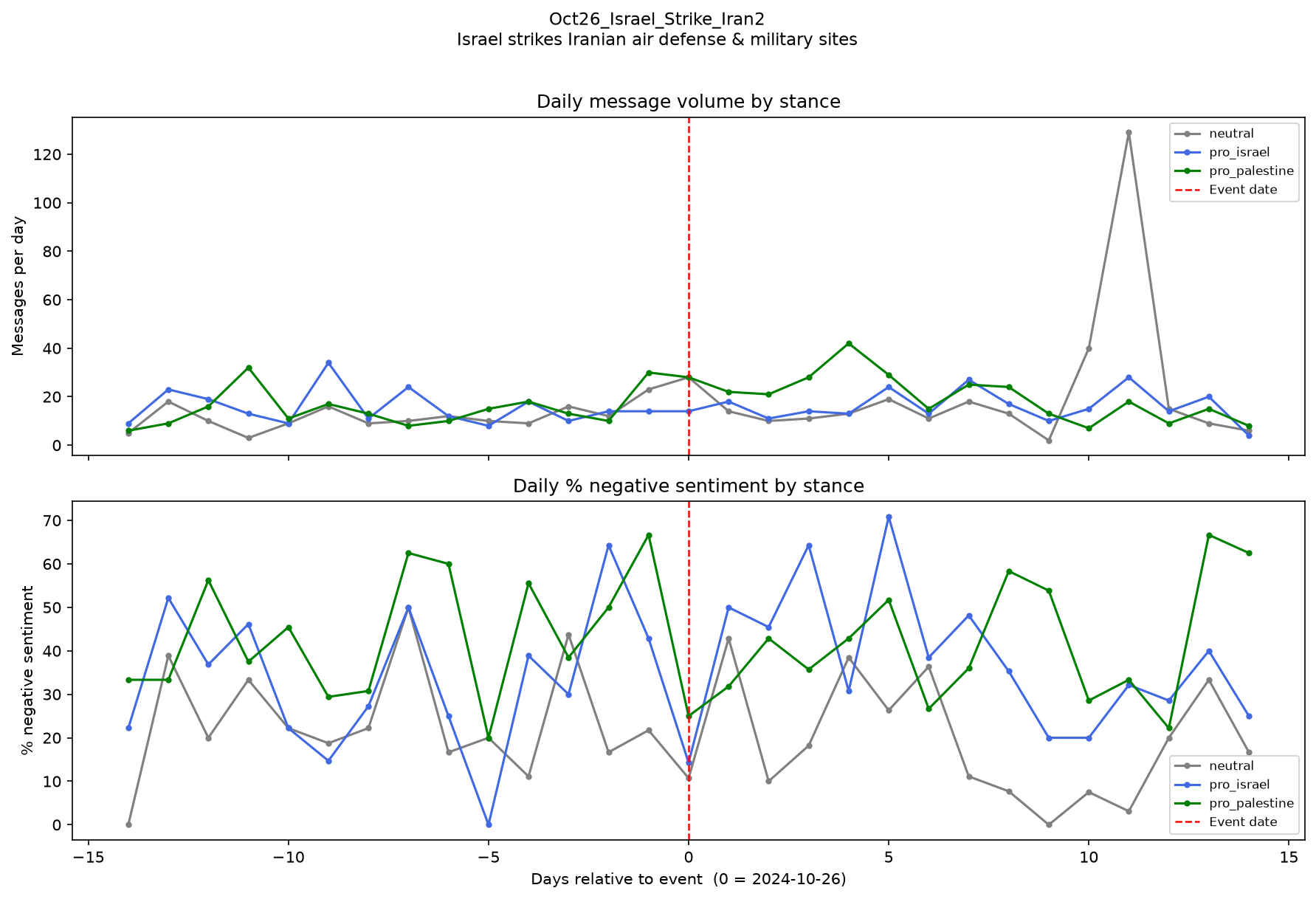}
\caption{Event window: Israel strikes Iranian air defense (25 Oct 2024).}
\label{fig:6_16}
\end{figure}
Israel Strikes Iranian Air Defense - October 25, 2024 On October 25, 2024, Israel struck Iranian air-defence systems and military infrastructure following the earlier Iranian missile attack. This event produced the weakest and least clearly structured response of the five. Day-0 volume was low across all groups (pro-Palestine 30, neutral 23, pro-Israel 14), and unlike the other military events there was no pronounced neutral spike at the event date; the largest neutral increase in this window occurs roughly twelve days later and most likely reflects a subsequent development bleeding into the window rather than a response to the strike itself. The ±14-day window of this event also overlaps the aftermath of the October 1 escalation: the pre-event baseline covers days 10 to 23 after the missile attack, so pre/post sentiment comparisons are confounded by construction. The sentiment record is consistent with this. The only clear movement, a pro-Israel rise from 35\% to 47\% in the first week, reverts fully by the second (34\%, back at its pre-event level), while pro-Palestine negativity is essentially flat throughout (42\%, 42\%, 45\%). An isolated, fully reverting shift on a contaminated baseline cannot be distinguished from residual movement of the October 1 aftermath, and no interpretation is drawn from it. The event is retained because it met the volume threshold applied to all five events and because its event-day composition contributes to the cross-event scaling pattern described below.

\begin{figure}[t]
\centering
\includegraphics[width=\linewidth]{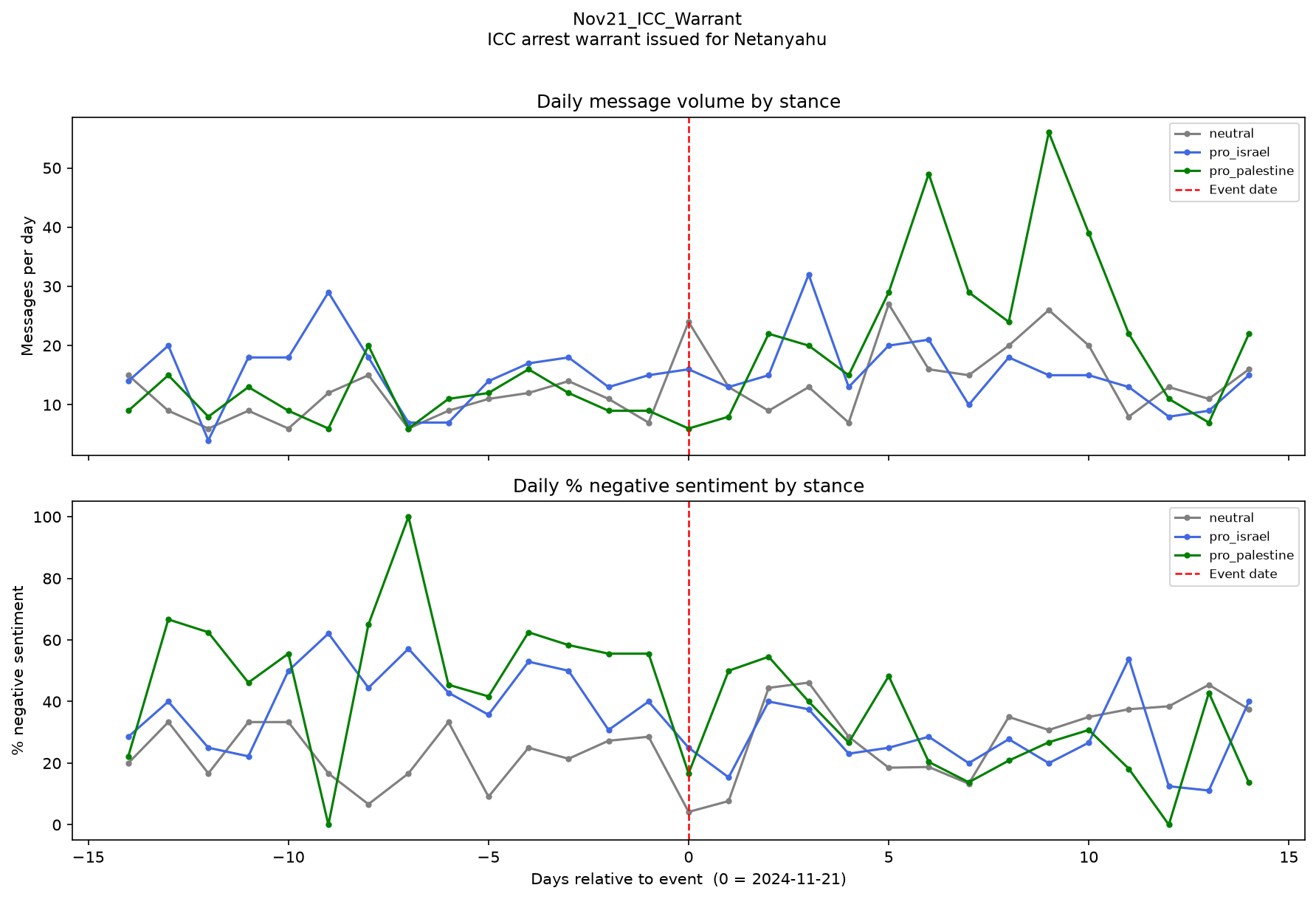}
\caption{Event window: ICC arrest warrant for Netanyahu (21 Nov 2024).}
\label{fig:6_17}
\end{figure}
ICC Arrest Warrant for Netanyahu - November 21, 2024 On November 21, 2024, the International Criminal Court issued an arrest warrant for Israeli Prime Minister Benjamin Netanyahu. This event behaved unlike the military escalations. Volume on the event day itself was low across all groups (neutral 24, pro-Israel 16, pro-Palestine 6); the main pro-Palestine increase came several days later, building as the legal decision was confirmed and political reactions accumulated. Negative sentiment in pro-Palestine messages fell substantially alongside this delayed volume increase, from 54\% in the two weeks before the warrant to 32\% in days 0--7 and 23\% in days 8--14, in contrast to military events, where heightened activity is typically accompanied by stronger negative sentiment. Pro-Israel negativity declined as well, from 43\% to 29\% in the first week and 29\% in the second, despite little change in volume. The warrant was discussed primarily as a legal and institutional development rather than an immediate escalation. Pro-Israel volume changed little, and no neutral spike of the kind seen during military events appeared, indicating that the event generated stance-based commentary rather than factual reporting.

\begin{figure}[t]
\centering
\includegraphics[width=\linewidth]{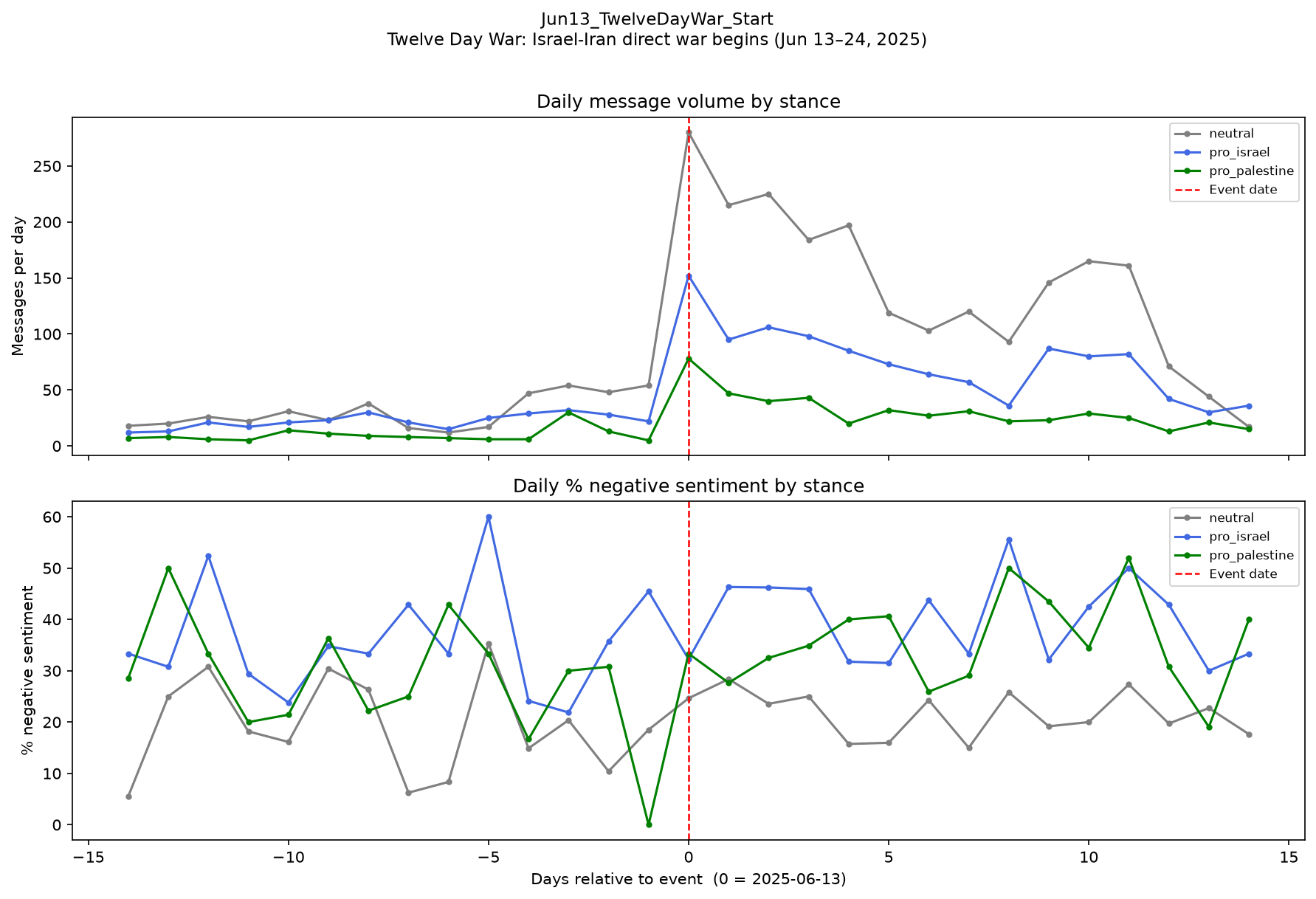}
\caption{Event window: Twelve Day War onset (13 Jun 2025).}
\label{fig:6_18}
\end{figure}
Twelve Day War - June 13, 2025 The onset of direct Israel--Iran hostilities on June 13, 2025 produced one of the largest volume increases in the dataset. All three stance labels rose simultaneously, with neutral reaching the highest absolute volume (280, against 152 pro-Israel and 78 pro-Palestine on the event day). As in other military events, pro-Israel channels produced a large share of neutral-labelled, report-style content during the initial reporting period. Pro-Israel negative sentiment was relatively low on the event day and rose over the following days as Iranian retaliation continued. Across the weekly windows the movement was upward on both sides but small: pro-Israel negativity moved from 36\% before the event to 39\% and 41\% in the two weeks after, and pro-Palestine from 29\% to 33\% and 39\%. After the initial escalation, the two groups' sentiment trajectories moved more closely together, both responding to the same sequence of developments while framing them differently.

\begin{figure}[t]
\centering
\includegraphics[width=\linewidth]{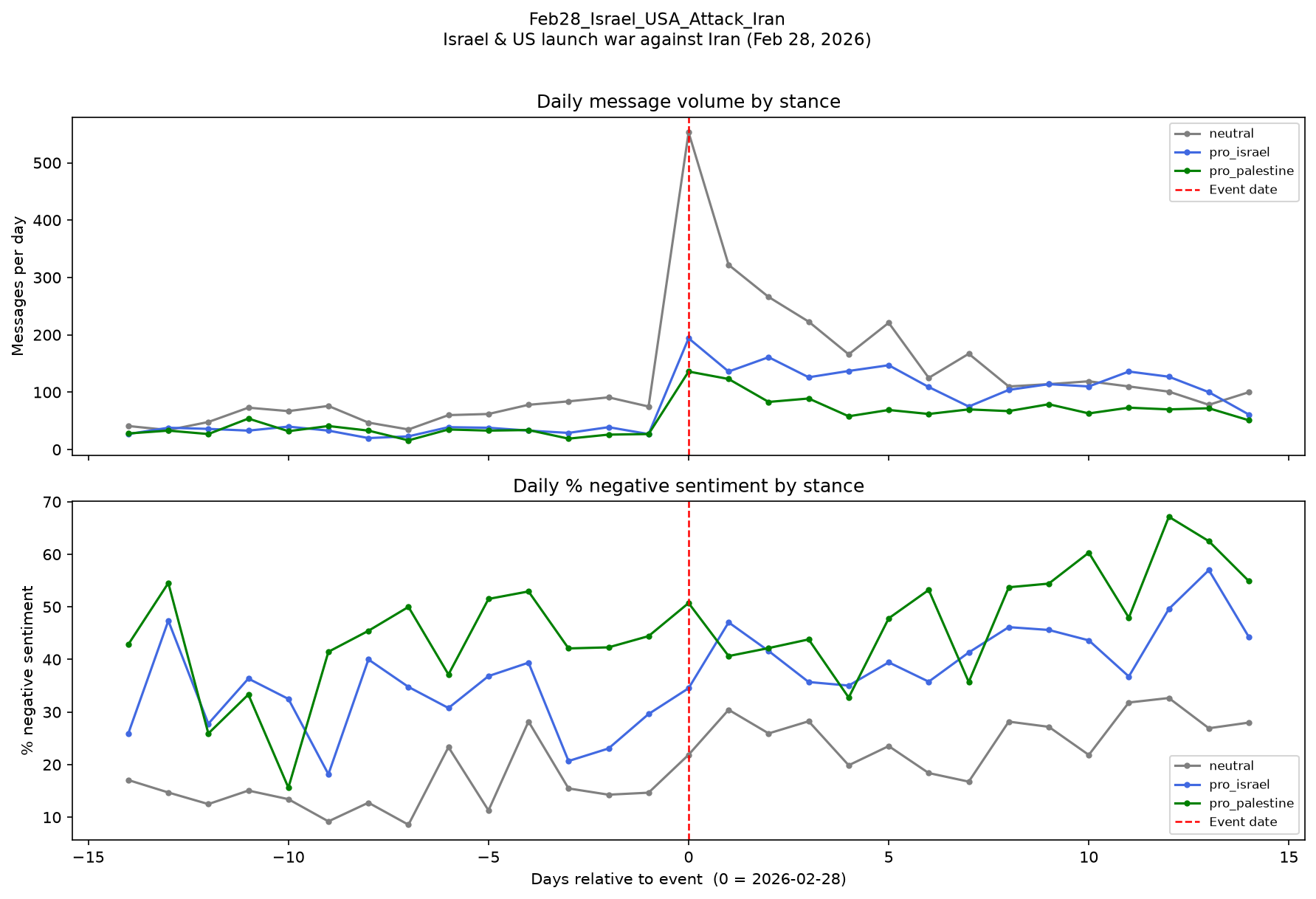}
\caption{Event window: Israel and US attack Iran (28 Feb 2026).}
\label{fig:6_19}
\end{figure}
Israel and US Attack Iran - February 28, 2026 The joint Israel--United States attack on Iran on February 28, 2026 produced the largest single-day volume increase in the dataset. All three stance categories rose, with neutral far exceeding both stance groups on the event day (553 messages, against 194 pro-Israel and 136 pro-Palestine). As all messages originate from politically aligned channels, this neutral surge does not represent a neutral community; it reflects a shift toward descriptive, news-style reporting, most pronounced in pro-Israel channels. Pro-Palestine channels showed stronger explicit stance expression, with proportionally fewer messages classified as neutral. The sentiment response was staggered but ultimately joint: pro-Israel negativity rose immediately, from 32\% before the attack to 39\% in the first week and 46\% in the second, while pro-Palestine negativity was initially little changed (41\% to 44\%) before rising sharply to 57\% in the second week as Iranian responses accumulated. This is the most recent escalation in the corpus. Because the underlying conflict was still unresolved at the close of the collection window, only a Memorandum of Understanding had been signed, not a lasting settlement, its full discursive aftermath extends beyond June 2026, and the trends reported here should be treated as preliminary. Figure~\ref{fig:6_19} Cross-event pattern Across the five events, event type shapes both the volume and the framing of the response. The three largest military escalations (Oct 1, Jun 13, Feb 28) produced large, immediate volume increases led by neutral and pro-Israel content, with the neutral rise concentrated in pro-Israel channels, a shift toward factual, report-style language rather than the presence of any neutral audience. This shift scales with the magnitude of the escalation: the neutral share of event-day volume rises monotonically with total event-day volume across all four military events, from 34\% on the smaller October 25 strike (23 of 67 messages) to 46\% on October 1, 55\% on June 13, and 63\% on February 28 (553 of 883). The ordering follows event size, not date, indicating that larger escalations push the channels further into report-style output rather than reflecting gradual change in the corpus over time. Pro-Palestine channels tended to maintain more explicit political framing throughout. The legal event (the ICC warrant) followed a distinct pattern: a delayed, stance-driven pro-Palestine increase with no neutral spike, indicating commentary rather than reporting. The October 25 strike, with too little volume for a clear response profile of its own, fits neither of these patterns cleanly, though its low event-day neutral share anchors the scaling relationship above. Overall, military developments are associated with increased neutral-labelled, report-style activity, especially in pro-Israel channels, while the legal development produced stronger explicit stance expression among pro-Palestine channels. The direction of the sentiment response depended on the structure of the event. During the two reciprocal escalations, negativity moved the same way in both communities: around February 28 it rose on both sides, immediately for pro-Israel (32\%, 39\%, 46\%) and by the second week for pro-Palestine (41\%, 44\%, 57\%), and around June 13 both sides shifted in the same upward direction, though only slightly. The one-sided strike of October 1 instead produced mirror-image movement: pro-Israel negativity rose from 30\% to 45\% in the first week while pro-Palestine negativity fell from 31\% to 22\% over the same days, before the two converged again in the second week. The ICC warrant produced the opposite of an escalation response: negativity fell in both groups, roughly twice as steeply on the pro-Palestine side (54\% before the warrant to 32\% and 23\% after, against 43\% to 29\% and 29\%), and for pro-Palestine channels the decline coincided with their delayed volume surge, the only event in which rising activity was accompanied by falling negativity, consistent with both sides discussing the ruling in an institutional register rather than through the casualty-centred vocabulary that drives negative classification. Each community's negative sentiment thus tracked the direction of events for its own side, extending the acting-party/affected-party asymmetry of Section~\ref{sec:framing-results} into the temporal domain: rising together when both sides were being struck, diverging when only one was, and subsiding in both groups when the development was institutional rather than kinetic.

\section{Discussion}
\label{sec:discussion}

\subsection{Model Performance and Method Selection}
\label{sec:disc-performance}
The three stance detection approaches produced clearly different results. Fine-tuned BERTweet reached 72.1\% accuracy and a macro F1 of 0.721, evaluated using 5-fold stratified cross-validation (F1 0.721 ± 0.027 across folds). The zero-shot DeBERTa model scored 64.5\% accuracy and 0.643 F1, and the keyword method 60.6\% accuracy and 0.608 F1. The fine-tuned model therefore outperformed the other two by roughly 8 to 11 percentage points on macro F1. What stands out is that keywords and zero-shot both landed well short of the fine-tuned model despite working in completely different ways. One relies on a hand-built list of conflict terms, while the other is a large transformer with no task-specific training. Neither had seen the manually annotated Telegram messages before evaluation, and neither cleared the mid-60s (60.6\% and 64.5\% accuracy), 8 to 11 points below fine-tuning. Telegram conflict channels mix political framing, regional jargon, informal language, and multi-actor references in ways a static lexicon cannot fully capture and a general-purpose NLI model has not been exposed to. Fine-tuning broke through that ceiling. BERTweet began from pre-training on 850 million tweets, a strong foundation for short, informal social-media text; adding 736 domain-specific labeled examples allowed it to learn the vocabulary patterns and framing conventions specific to this dataset. For similar work, stance detection in politically polarised Telegram content, the results suggest fine-tuning on even a small amount of in-domain labeled data can outperform both lexicon-based and zero-shot approaches. Because performance was measured with cross-validation over the full annotated set rather than a single held-out split, the reported figure (0.721 ± 0.027) reflects a stable estimate rather than the outcome of one fortunate partition. The inter-method agreement results reinforce this from a different angle. Where all three methods assigned the same label, the message likely contained clear, explicit stance markers any approach could detect. Where they disagreed, roughly 40\% of the dataset, stance was expressed through framing, context, or indirect language that a lexicon missed, a general-purpose NLI model did not recognise, and only the fine-tuned model partially resolved. The disagreement rate is therefore not a verdict on which method is better; it measures how much of the dataset relies on implicit rather than explicit stance expression. That disagreements almost always involved one method defaulting to neutral, rather than two methods disagreeing on which side was meant, confirms this: the methods rarely contradicted each other on political direction, they differed on whether a political signal was present at all.

\subsection{Interpretation of Key Findings}
\label{sec:disc-interpretation}
Across all three detection methods, messages with an explicit political stance carried substantially more negative sentiment than messages classified as neutral. The pattern holds regardless of which method assigned the label, so it cannot be attributed to any particular classification approach. Politically aligned Telegram channels, at least in this dataset, pair political positioning with negative emotional language consistently. The framing results add something more specific. Pro-Israel channels used death-context and victim framing more often than pro-Palestine channels, 27.8\% versus 17.7\% for death context, 19.8\% versus 11.9\% for victim framing. On the surface that looks counterintuitive. But the sentiment breakdown changes the picture entirely. In pro-Israel messages flagged for death context, sentiment is predominantly neutral (around 54\%, against 45\% negative); in pro-Palestine messages with the same framing, negative sentiment reaches roughly 62\%. The same vocabulary, a completely different emotional register. The two sides are not talking about different topics. They talk about the same events from different positions. Pro-Israel channels more often write as an actor, reporting operations, documenting what happened, framing military decisions, so death appears as the reported outcome of an action rather than a loss suffered, and a factual register follows. Pro-Palestine channels more often write as an affected party: the same deaths, the same destruction, but described from inside the experience. These differences surface in the classifier's sentiment labels even where the surface vocabulary overlaps, though sentiment was not separately validated against human annotation on this corpus. That gap has practical implications for how conflict discourse is studied. A method that looks only at sentiment misses cases where high death-context vocabulary and neutral sentiment coexist; a method that looks only at stance misses the emotional texture separating two messages that make the same claim. Neither dimension alone is sufficient.

\subsection{Limitations}
\label{sec:limitations}
The dataset covers 16 Telegram channels, 8 pro-Palestine and 8 pro-Israel. That is a narrow slice of a much larger information environment. The reverse-chronological collection method means earlier time periods are underrepresented; most messages in the dataset were published after October 2023, which limits how much can be said about the pre-October 7 period. A larger and more temporally balanced dataset would support stronger longitudinal claims. The keyword lexicons were built by one person and every decision about which terms to include, which side to assign them to, or how to weigh them reflects individual judgment. The lexicons were developed through years of familiarity with the channels and the conflict, but they were never validated by a second annotator, so there is no objective way to verify that the weighting choices are correct. For messages that mix terminology from both sides, or that discuss the Iran-related escalations using vocabulary outside the original Gaza-focused scope, the keyword method is likely to underperform. A further consideration is the size asymmetry between the two lexicons: the pro-Palestine lexicon contains 227 terms compared to 169 for the pro-Israel lexicon. More surface area for matching produces systematically higher recall for the larger lexicon, all else equal. This is consistent with the evaluation results in Section~\ref{sec:eval-keyword}, where the keyword method's pro-Palestine recall (\textasciitilde{}62\%) exceeded its pro-Israel recall (\textasciitilde{}53\%). Interpreting keyword-based stance distribution results across the full dataset should account for this asymmetry. The annotation set has the same single-annotator limitation. Seven hundred and thirty-six messages is a small ground truth, and every label was assigned by one person. In subjective classification tasks, agreement is consistently lower for ambiguous categories than for clearly separable ones, because the ambiguity of the data and the task directly depresses the agreement coders reach \cite{nowak2010}. The neutral class is most exposed to this, because the boundary between neutral and politically aligned is genuinely unclear in many messages \cite{kucuk2020}; and since inter-annotator agreement could not be calculated with a single annotator, the reliability of these borderline labels cannot be verified. All three methods were evaluated on the same 736 annotated messages, making the comparison directly fair. For the fine-tuned model, evaluation used out-of-fold predictions from 5-fold cross-validation: each message was classified by a fold that had not seen it during training, which removes train--test leakage while still allowing the full annotated set to be used for evaluation. This is a stronger basis for comparison than a single held-out split, particularly given the limited size of the annotated set. Critically, although the reliability of the ground-truth labels themselves cannot be quantified without a second annotator, this limitation applies equally to all three methods. Because every method is evaluated against the identical 736-message label set, any annotation noise enters all three evaluations the same way and cannot, on its own, produce the systematic 8-to-11-point gap by which the fine-tuned model outperforms both baselines. The single-annotator constraint therefore bounds the interpretation of the absolute scores, not the relative ranking of the methods. The dataset was limited to channels with a clear and consistent political orientation, either pro-Palestine advocacy channels or Israeli news and commentary outlets covering the conflict from an Israeli perspective. That was a methodological choice, not an oversight; the goal was to study how opposing communities communicate, not to map the full range of conflict discourse. As a consequence, the findings do not extend to general news channels, multilingual content, or communities that sit somewhere between the two poles and the results should be interpreted within that scope. A final limitation concerns the sentiment component: unlike the stance model, the Twitter-RoBERTa sentiment classifier was applied off-the-shelf, without fine-tuning or validation against human-annotated sentiment on this corpus. Because the central register contrast of Section~\ref{sec:framing-results} is expressed through these sentiment labels, the specific proportions should be treated as model-derived estimates rather than validated measurements, though the magnitude and consistency of the observed gaps make a reversal of the finding's direction unlikely.

\subsection{Broader Context}
\label{sec:broader}
The channels in this dataset operate as closed information environments. Each one frames events consistently from one political position, and subscribers self-select into those streams. This aligns with research on echo chambers in social media, where users tend to cluster around content that reinforces existing beliefs rather than being exposed to opposing viewpoints \cite{cinelli2021}, \cite{baumann2020}. The sentiment--stance pattern observed here can be related to a parallel study of the same conflict across Telegram, Reddit, and Twitter. Antonakaki and Ioannidis found that anger and fear were the dominant emotions, particularly in channels focused on direct conflict reporting, with highly polarised emotional responses \cite{antonakaki2025telegram}. The present study measures sentiment polarity rather than discrete emotions, and finds neutral sentiment dominant across the dataset as a whole, a reflection of the large volume of factual, report-style content. Within the politically engaged subset, however, the two findings align: messages expressing an explicit stance consistently show elevated negative sentiment, echoing the anger- and fear-driven polarization reported in \cite{antonakaki2025telegram}. A similar pattern appears in earlier analysis of the 2014 Gaza war on Twitter: Siapera, Hunt, and Lynn \cite{siapera2015} found conflict discourse dominated by negative sentiment built around vocabulary of casualties, children, and hospitals, alongside a distinct humanitarian register that cut across the pro-Israel and pro-Palestine poles \cite{siapera2015}. The framing asymmetry identified in the present corpus refines that picture, showing that the two groups draw on the same death- and victim-centred vocabulary while charging it with sharply different sentiment.

\section{Conclusion}
\label{sec:conclusion}
This study set out to study how politically aligned Telegram channels discuss the Israel-Palestine conflict, using a combination of sentiment analysis, stance detection, and framing analysis. The final dataset contained 87,617 messages collected from 16 channels, 8 pro-Palestine and 8 pro-Israel, spanning May 2021 to June 2026. Three stance detection methods were built and evaluated: a keyword-based approach, a zero-shot classifier using DeBERTa, and a fine-tuned BERTweet model trained on 736 manually labeled messages. Sentiment was analysed across the full dataset using a Twitter-RoBERTa model, while framing was examined through four keyword-defined categories tied to death, victims, military operations, and legal accountability. On performance, fine-tuned BERTweet was the clear winner at 72.1\% accuracy and 0.721 macro F1, roughly 8 to 11 points above both the keyword method and zero-shot DeBERTa, which clustered in the low-to-mid 60s (60.6\% and 64.5\%). This figure was obtained through 5-fold cross-validation over the full annotated set (0.721 ± 0.027), so it reflects a stable estimate rather than a single fortunate split. The fact that two completely different methods both stalled well below the fine-tuned model is itself a finding: without adaptation to the specific language of this dataset, there is a hard limit on what lexicon-based and zero-shot stance detection can achieve here. Messages classified as pro-Palestine or pro-Israel, regardless of which detection method assigned the label, carried far more negative sentiment than neutral-classified ones. That held across all three methods. The most interesting result came from framing. Pro-Israel channels used death and victim vocabulary more often than pro-Palestine channels, but with predominantly neutral sentiment; pro-Palestine channels covered the same topics with substantially higher negative sentiment. The two sides occupy different narrative positions: one writes more often as a reporting actor, the other more often as an affected party. This interpretation follows from the combination of framing-vocabulary rates and sentiment register rather than from a separately validated frame analysis, and it becomes visible only when stance, sentiment, and framing are examined together rather than in isolation. The event analysis extended this asymmetry into the temporal domain. Across five documented conflict events, volume and framing responded to event type: military escalations produced immediate, report-style surges concentrated in pro-Israel channels, while the ICC warrant produced a delayed, stance-driven pro-Palestine response. Each community's negative sentiment tracked the direction of events for its own side, rising in both groups during reciprocal escalations, diverging after the one-sided October 1 strike, and falling in both groups after the ICC ruling. Several directions remain open for future work. The dataset is English-only, but a large share of conflict-related Telegram content is in Arabic; extending the analysis to multilingual content would give a considerably more complete picture of how the conflict is discussed across communities. Cross-platform comparison is another obvious next step; Twitter and Reddit would be natural candidates, though both platforms currently restrict API access in ways that make large-scale collection difficult. On the methodology side, a larger annotated dataset built by multiple annotators would strengthen the evaluation and likely push fine-tuned performance higher, since a single annotator working through 736 messages is a starting point, not a ceiling. A further methodological extension would be to apply the same comparative approach used for stance detection to the sentiment component: the present study used a single pre-trained model for sentiment classification, but future work could fine-tune a domain-specific sentiment model on manually annotated conflict-related messages and compare it against general-purpose pre-trained models, mirroring the multi-method comparison conducted here for stance. This would establish whether the domain-specificity that proved advantageous for stance detection yields similar gains for sentiment. Finally, bot detection and coordinated-behaviour analysis within Telegram channels was not explored here but could add a useful layer, understanding not just what these communities say, but how organised or automated that activity is.

\section{Acknowledgements}
\label{sec:acknowledgements}
I would like to thank my supervisor, Prof. Sotiris Ioannidis, for the opportunity to carry out this study under his supervision. My deepest thanks go to Despoina Antonakaki, whose guidance shaped this work from beginning to end. She helped me develop the direction of the study, offered ideas at every stage, and provided the support and feedback that made it possible to complete. I am genuinely grateful for her time and patience. I am profoundly thankful to my family, and especially my parents, for their patience and unwavering support throughout my studies. I know how much they wanted this for me, and reaching this point is as much theirs as it is mine. To my closest friends, Achilleas, Apostolou, Koukas, Orestis, Thomas, and Tsompanikos, I owe you more than I can put into words. You are family to me, every bit as much as my own, and the years behind this study would have meant nothing without you beside me. Thank you for all of it. I also want to thank my pets who gave me more than they will ever know. My dogs Luffy and Charlie. And my cats: Irida, who is no longer with me but who once gave me a reason to keep going when I needed one most; Adis; and Mahsa, who keeps me company now.

\bibliographystyle{plainnat}
\bibliography{paper}

\end{document}